\pdfoutput=1

\documentclass[11pt]{article}

\usepackage[final]{acl}

\usepackage{times}
\usepackage{latexsym}

\usepackage[T1]{fontenc}

\usepackage[utf8]{inputenc}

\usepackage{microtype}

\usepackage{inconsolata}

\usepackage{xcolor}
\usepackage{hyperref}
 \definecolor{darkblue}{rgb}{0, 0, 0.5}
  \hypersetup{colorlinks=true, citecolor=darkblue, linkcolor=darkblue, urlcolor=darkblue}

\usepackage{xstring}

\usepackage{color}

\usepackage{booktabs}
\usepackage{multirow}
\usepackage{tikz}
\usepackage{algorithmic}
\usepackage{algorithm}
\usepackage{listings}
\usepackage{xcolor}
\usepackage{amsmath}

\definecolor{codegreen}{rgb}{0,0.6,0}
\definecolor{codegray}{rgb}{0.5,0.5,0.5}
\definecolor{codepurple}{rgb}{0.58,0,0.82}
\definecolor{backcolour}{rgb}{0.95,0.95,0.92}

\lstdefinestyle{mystyle}{
    backgroundcolor=\color{backcolour},   
    commentstyle=\color{codegreen},
    keywordstyle=\color{magenta},
    numberstyle=\tiny\color{codegray},
    basicstyle=\ttfamily\footnotesize,
    breakatwhitespace=false,         
    breaklines=true,                 
    captionpos=b,                    
    keepspaces=true,                 
    numbers=left,                    
    numbersep=5pt,                  
    showspaces=false,                
    showstringspaces=false,
    showtabs=false,                  
    tabsize=2
}

\usepackage[colorinlistoftodos,prependcaption,textsize=tiny]{todonotes}

\title{Generating Signed Language Instructions in Large-Scale Dialogue Systems}

\author{
    Mert İnan\textsuperscript{\rm 1}, 
    Katherine Atwell\textsuperscript{\rm 1}, 
    Anthony Sicilia\textsuperscript{\rm 1},
    Lorna Quandt\textsuperscript{\rm 2},
    Malihe Alikhani\textsuperscript{\rm 1}\\
    \textsuperscript{\rm 1} Khoury College of Computer Science, Northeastern University, Boston, MA, USA \\
    \textsuperscript{\rm 2} Educational Neuroscience Program, Gallaudet University, Washington, D.C., USA\\
    \texttt{\{inan.m, atwell.ka, sicilia.a, alikhani.m\}@northeastern.edu} \\
    \texttt{lorna.quandt@gallaudet.edu} 
}

\begin{document}
\maketitle
\begin{abstract}
We introduce a goal-oriented conversational AI system enhanced with American Sign Language (ASL) instructions, presenting the first implementation of such a system on a worldwide multimodal conversational AI platform. Accessible through a touch-based interface, our system receives input from users and seamlessly generates ASL instructions by leveraging retrieval methods and cognitively based gloss translations. Central to our design is a sign translation module powered by Large Language Models, alongside a token-based video retrieval system for delivering instructional content from recipes and wikiHow guides. Our development process is deeply rooted in a commitment to community engagement, incorporating insights from the Deaf and Hard-of-Hearing community, as well as experts in cognitive and ASL learning sciences. The effectiveness of our signing instructions is validated by user feedback, achieving ratings on par with those of the system in its non-signing variant. Additionally, our system demonstrates exceptional performance in retrieval accuracy and text-generation quality, measured by metrics such as BERTScore. We have made our codebase and datasets publicly accessible at \url{https://github.com/Merterm/signed-dialogue}, and a demo of our signed instruction video retrieval system is available at \url{https://huggingface.co/spaces/merterm/signed-instructions}.



\end{abstract}

\begin{figure}[t!]
    \centering
    \includegraphics[width=.9\columnwidth]{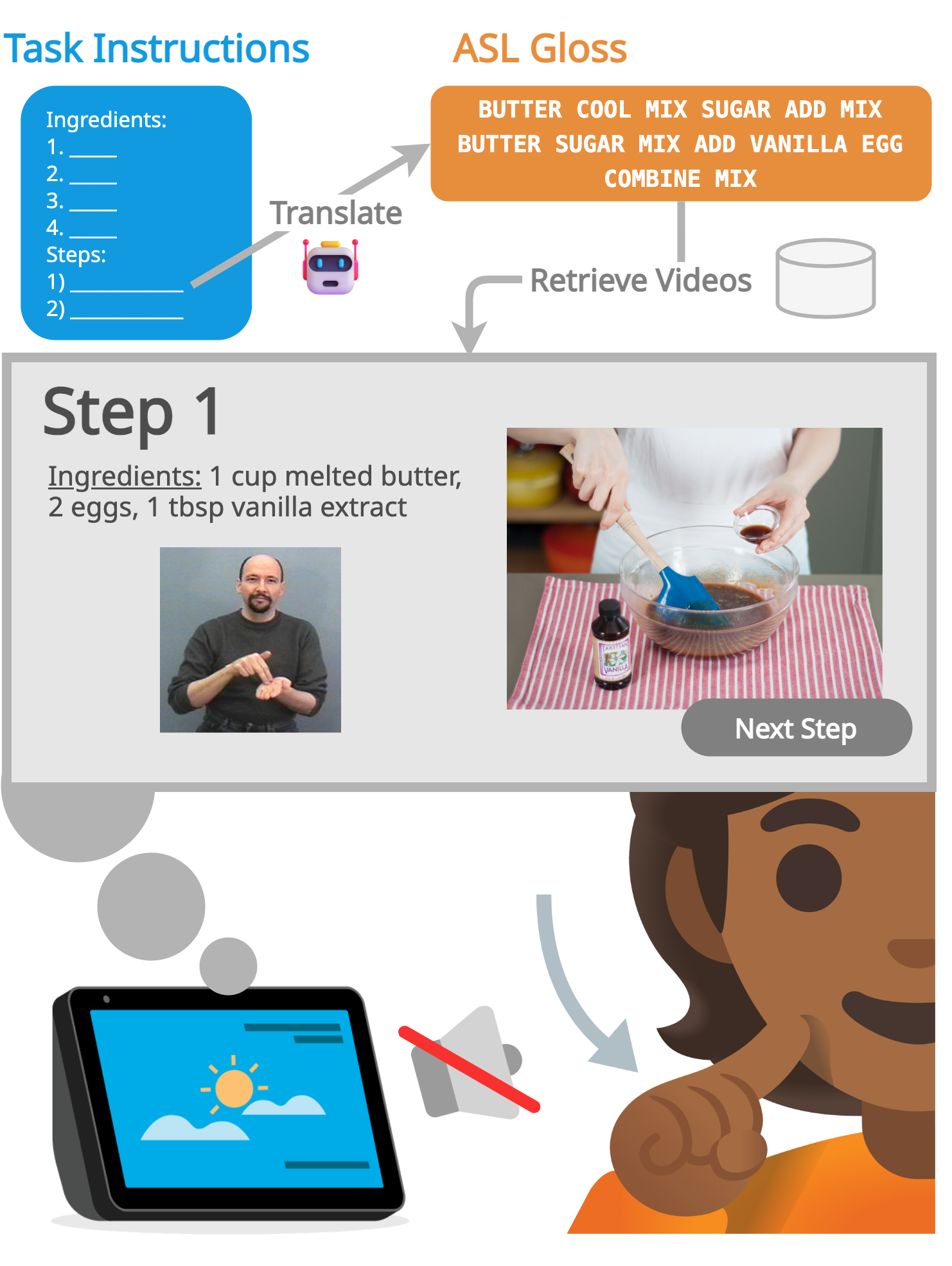}
    \caption{An overview of our multimodal dialogue system, capable of giving signed instructions to Deaf or Hard-of-Hearing users in ASL. We first translate task instructions to an intermediate textual representation called glosses using Large Language Models; then, we fetch token-level sign videos to display on the screens of Amazon Alexa Echo Show. 
    }
    \label{fig:sign_pipeline}
\end{figure}

\section{Introduction}
Conversational systems have become increasingly integrated into our everyday lives, yet their accessibility to the Deaf and Hard-of-Hearing (DHH) community, who predominantly communicate through signed languages, remains limited \cite{Glasser2017Oct, Glasser2020Jul, Bragg2020Apr}. Despite growing advocacy for more inclusive interactive technologies from DHH users \cite{Bragg2019Oct, Blair2020Dec, Kahlon2023Mar}, a comprehensive dialogue system tailored for sign language users has yet to be implemented on a global scale. In response, within the Alexa Prize TaskBot Challenge 2 framework, we developed and launched the first task-oriented, multimodal dialogue system utilizing ASL, aiming to bridge the gap between DHH users and personal voice assistants. This system translates touch-based inputs into ASL video instructions, offering a groundbreaking approach to interaction. This paper introduces our ASL instruction framework, marking a significant stride towards integrating conversational systems into the living spaces of sign language users and enhancing accessibility for the DHH community.


Many signers prefer to use ASL instead of text due to grammatical and linguistic differences between spoken and signed languages \cite{Hariharan2018Jun, Dangsaart2008Nov}. Yet currently, systems claiming to be accessible resort to text-based communication. As an alternative, videos or avatars of signers are options, yet these technologies are underutilized. In this paper, we show that deploying these signed systems on a large scale is, in fact, possible without much production cost and makes the system accessible to DHH users.



\begin{figure}[!h]
    \centering
    \begin{tabular}{c}
        \includegraphics[width=.85\columnwidth]{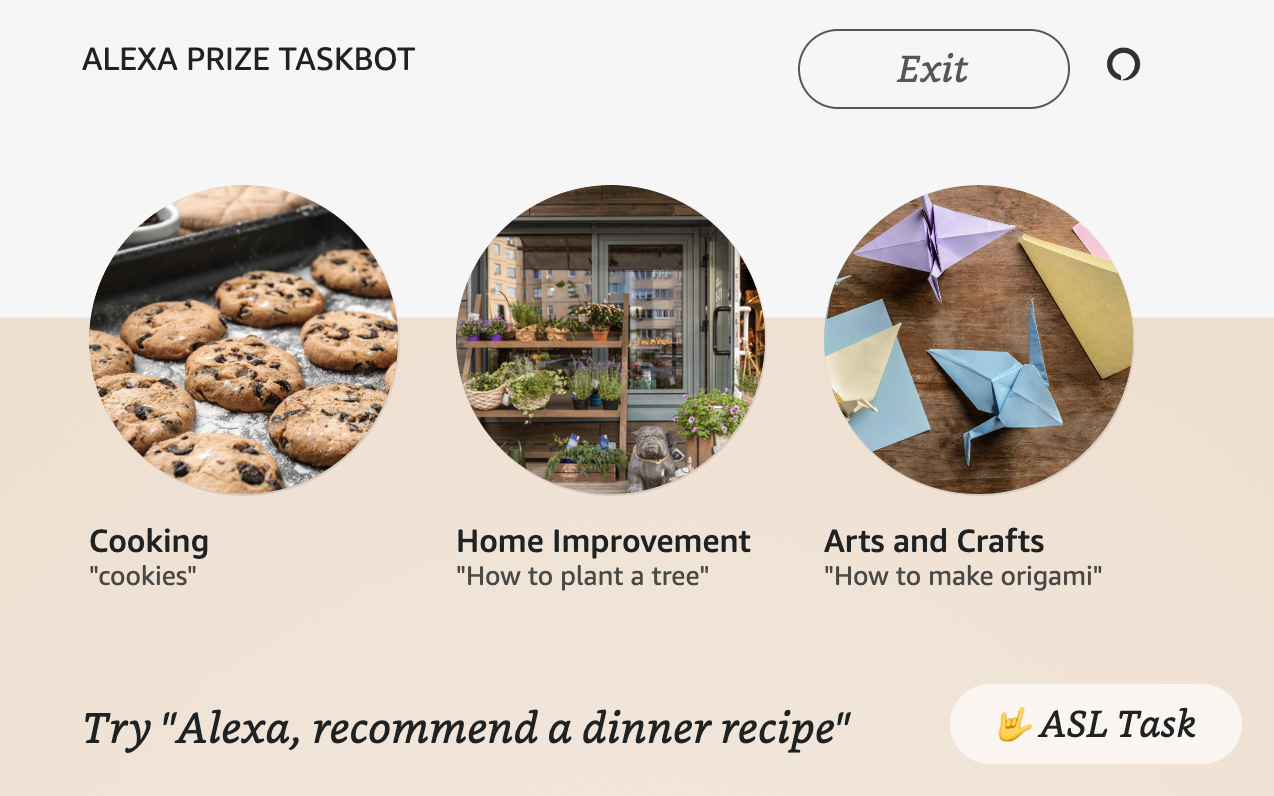} \\
        \includegraphics[width=.85\columnwidth]{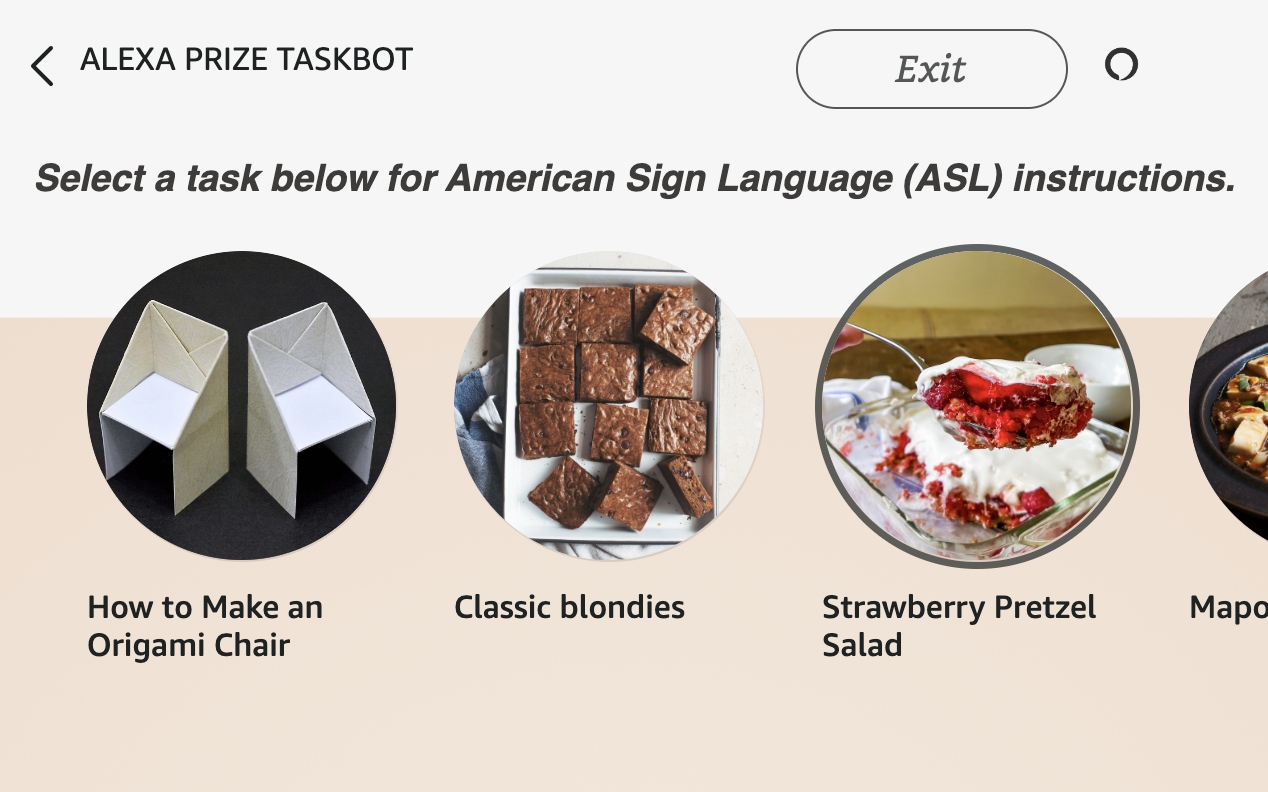} \\ \includegraphics[width=.85\columnwidth]{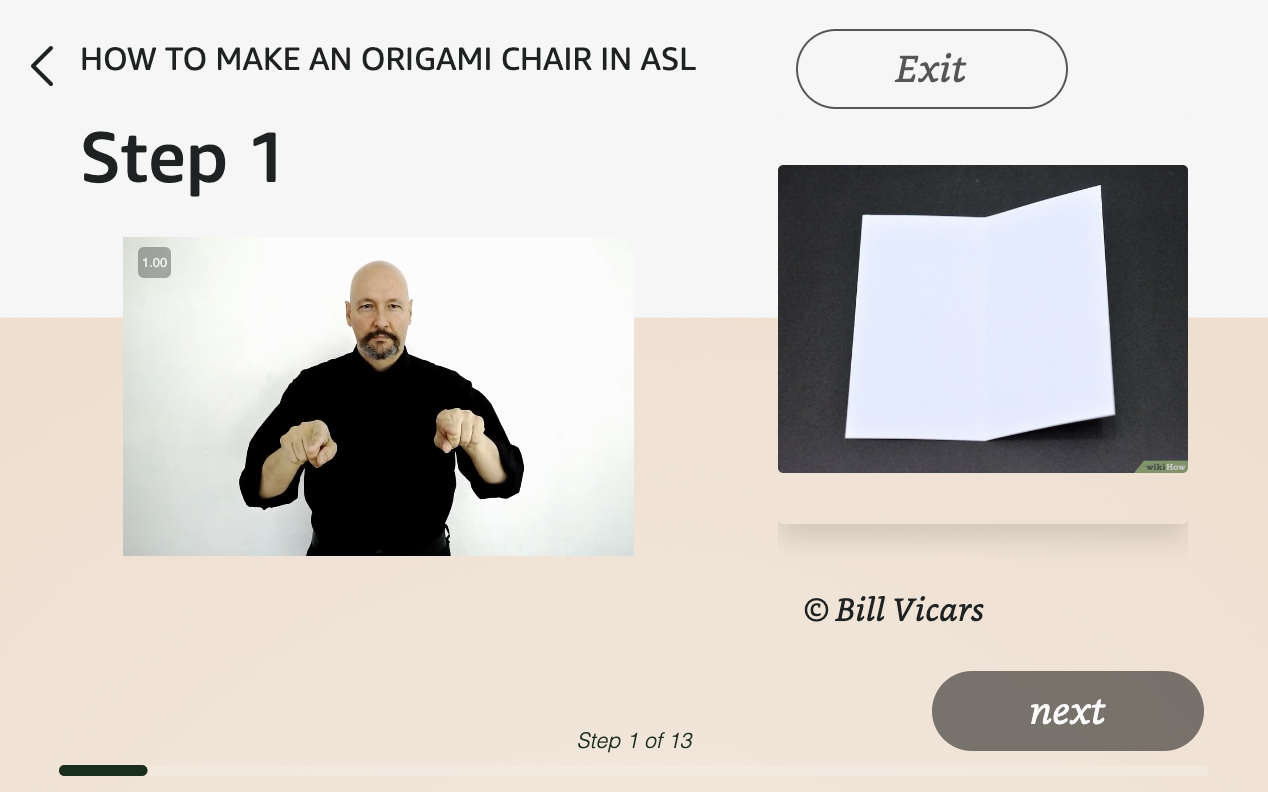} \\
        \includegraphics[width=.85\columnwidth]{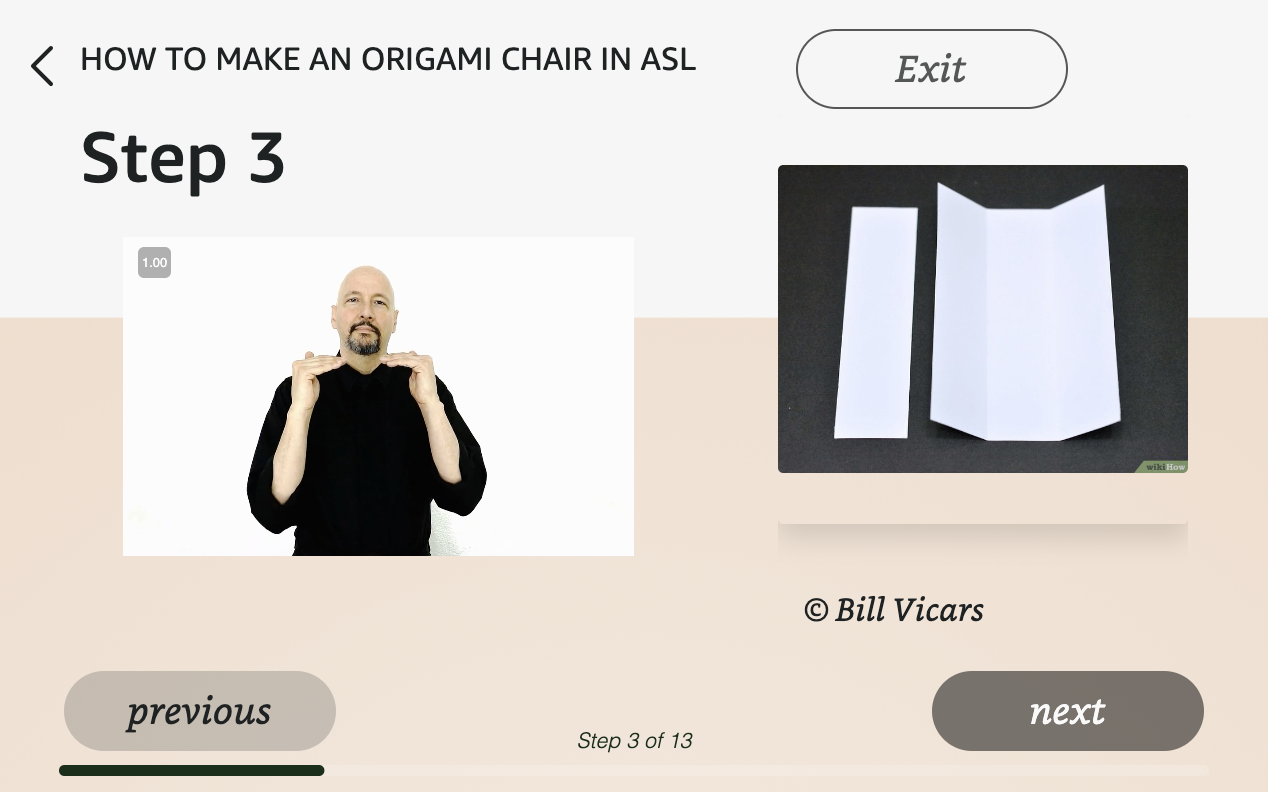} 
    \end{tabular}
    \caption{A storyboard of all the screens for an origami task with ASL video instructions. The first screen from the top is the landing page with an ASL Task button to enter the signed section. The second screen shows different recipes and task options. The following screens show an instruction step. Button interactions are especially important for signers as the audio is inaccessible.}
    \label{fig:sign_demo}
\end{figure}

Further, prior linguistics research has shown that DHH community members can experience higher cognitive loads while reading compared to signing \cite{Traxler2000Sep, kelly2003considerations, luckner2008summary}. In this paper, we investigate effective strategies of multimodal information presentation for the DHH to reduce cognitive load. With repeated consultations with cognitive scientists, we design the layout of our system's user interface specifically around the cognitive load of signers (see Figure~\ref{fig:sign_demo}).

We focus on creating a framework that is applicable to a large-scale global platform (in our case, Amazon Alexa), making it impossible at this time to access camera footage. We investigate ways of receiving input with other modalities instead of voice commands and without camera access. This leads us to focus on the task of instruction generation and delivery rather than recognizing signs produced by the user. We receive input from the user via touchscreen controls of Amazon Alexa Echo Show devices so that signers can interact without using voice commands (see Figure~\ref{fig:sign_demo} for the touch screen user interfaces where the user can interact via buttons to select tasks and navigate instructions). 

To address all of the aforementioned points, in the following sections, we introduce the components of our framework. Our detailed contributions are as follows:
\begin{enumerate}
    \setlength\itemsep{-0.1em}
    \item We design a multimodal task-oriented dialogue system with signed instructions and deploy it on multimodal devices.
    \item We use \textit{co-design} to build our system, actively involving community members in the design, development, and evaluation, ensuring our solutions positively impact the community. 
    \item We implement a novel Large Language Model (LLM)-based instruction generation technique for zero-shot text-to-sign translation. We use linguistics rules and cognitive science-based heuristics for this translation. 
    \item We make available a standalone library to translate instruction texts into signed instruction videos, and we release our dataset used for the top 200 signs in cooking and wikiHow domains.
\end{enumerate}

We hope this effort brings more focus to the needs of signers and will be a step towards making large-scale dialogue systems more accessible to all users.
\section{Related Work}
With the rise of voice assistant devices, the DHH community has been mostly left behind. Yet, there have been multiple lines of work to make them more accessible. Accessibility of personal assistant devices to the Deaf and Hard of Hearing community has been assessed multiple times before by \citet{Glasser2017Oct, Glasser2020Jul, Bragg2020Apr}. In addition, design approaches incorporating the DHH community have been proposed by \citet{Anindhita2016Aug, Hariharan2018Jun}. We build on these in our system design.

Most of the current work in interactive system design focuses on sign recognition with the help of cameras. For instance, in \citet{Wojtanowski2020} Wizard-of-Oz studies have been done where Alexa is combined with a camera to detect signs. In SIGNS project\footnote{\url{https://projectsigns.org/}}, Alexa recognizes specific gestures for simple task completion (such as getting the weather forecast with a specific gesture), and \citet{Huang} recognized signs for a healing robot. Even though these systems provide a means for recognizing signs, they fall short in generating signs, which we focus on in this paper.  

There has been some line of work by \citet{NasihatiGilani2019Jul} in generating avatars for 6-month-old babies to learn ASL. Also, \citet{Hruz2011Oct} deployed a kiosk with sign recognition and generation capabilities for Czech Sign Language. However, these have not resulted in a widely available system.

\begin{figure*}[!h]
    \centering
    \includegraphics[width=.75\textwidth]{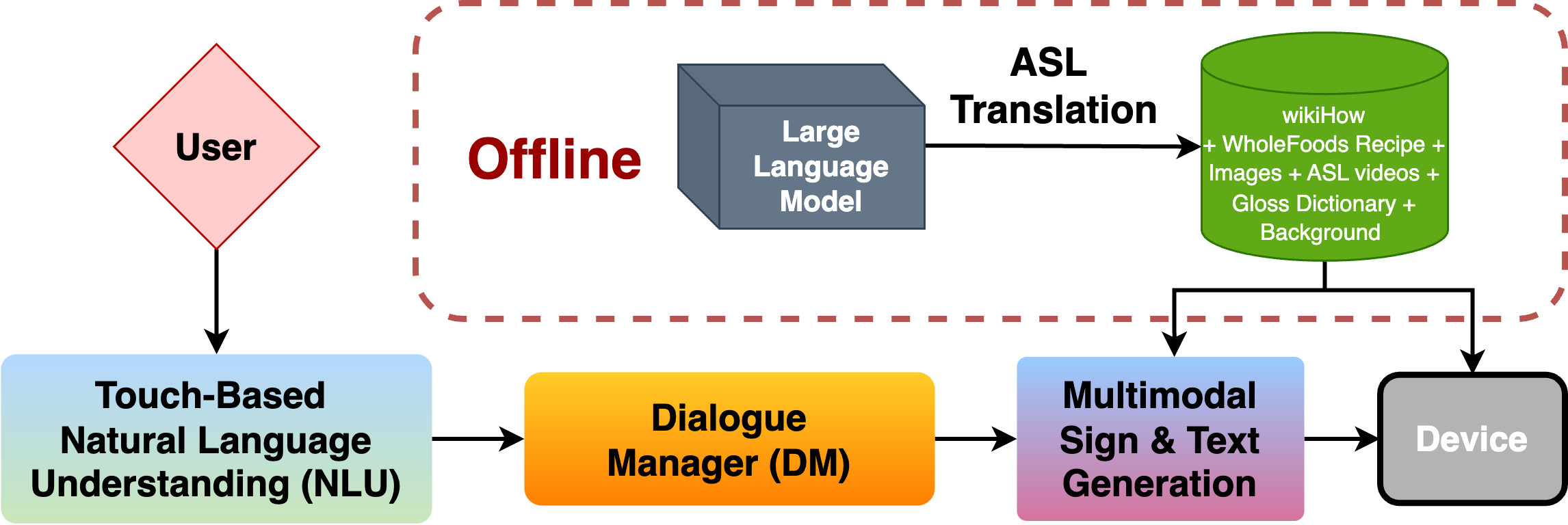}
    \caption{The overall architecture of our dialogue system with sign instructions for American Sign Language. Offline LLM translations make it easier to plug in a signing module into a traditional dialogue architecture.}
    \label{fig:dialogue_sys}
\end{figure*}

On the other hand, sign language processing has been widely studied under controlled conditions. Even though sign language generation and translation tasks are still open problems, transformer-based models in \citet{yin-read-2020-better, yin-etal-2021-including, moryossef-etal-2021-data, inan-etal-2022-modeling, muller-etal-2023-considerations, lin-etal-2023-gloss, viegas-etal-2023-including} have shown that it is possible to automate them better. As a core contribution, we present a framework to apply any of these models in large-scale interactive environments.


In order to make our system useful for signers, we need to mitigate their cognitive load interpreting instructions from multimodal devices. Models for the cognitive aptitudes and cognitive loads of sign language interpreters have been studied before by \citet{Macnamara2012, DuToit2017, Tiselius2018Jun, Chambers2020}. These models help guide the design principles of our system, as the user will need to focus on multiple modalities simultaneously through the visual modality, which increases cognitive load.

\section{A Goal-Oriented Dialogue System with Signed Instructions}
We design a multimodal goal-oriented dialogue system as part of the Alexa Prize TaskBot Challenge 2 \cite{TaskBot2023} and incorporate signed instructions. The main dialogue system that we develop follows a typical modular design: Natural Language Understanding (NLU), Dialogue Manager (DM), and Natural Language Generation (NLG). In this setting, we embed signed instructions into the multimodal NLG module (Figure~\ref{fig:dialogue_sys}). 

Due to privacy regulations, Alexa does not allow third parties to process user gestures and videos. Hence, to increase accessibility for signers, we choose to generate signed instructions instead of recognizing signs. To support users who cannot---or prefer not to---provide voice input, our system has a scrollable touchscreen with buttons. This enables us to have a full dialogue system for signers while complying with regulations.  

\subsection{Task Description}
We take as input a task JSON with step-by-step English text instructions, images, title, main image, and ingredients and output a JSON array of user interface screens corresponding to the gloss translations for each step and their corresponding sign videos (see Appendix~\ref{sec:inputs}). The tasks are in the domains of cooking, home improvement, arts and crafts, and gardening. We provide our signed instruction generation as a standalone library for the camera-ready version of this paper. 

\subsection{Community Co-Design}
\label{sec:signers}

To inform our system design choices, we connect with collaborators from the Deaf and Hard of Hearing (DHH) signing community at Gallaudet University (a prestigious higher education institution chartered for the DHH community). We incorporate the feedback from signers into the system's design. 


The feedback incorporated into our design process includes considering the cognitive load of signers, altering the dimensions of the text, video, and images used to communicate instructions, choosing which information to present as text versus signed videos (compare screens in Figure~\ref{fig:sign_demo} and Appendix~\ref{sec:interface} for the placement of text and signed videos in the same screen), and updating the design of the interface for ASL signers.
\section{Our Signed Instruction Framework}
We employ the framework shown in Figure~\ref{fig:sign_pipeline} to generate signed instructions. We first retrieve instructions for a given task, and then we convert each step into gloss 
tokens, which are intermediary textual representations using rule-based sign language translation algorithms and LLMs. Afterward, we segment each instruction into separate gloss tokens, retrieve sign videos for each, and stitch them back-to-back to create a continuous video sequence. For each step, we display this sequence of videos and a picture of the step. The picture for each step generally shows the result of the action as described in the sign instructions. This approach is summarized in Algorithm~\ref{alg:sign}.

\begin{algorithm}[!h]
\begin{algorithmic}[1]
\caption{Signed Instruction Retrieval}\label{alg:sign} 
\STATE $G \gets \{\}$ 
\STATE $I \gets$ instruction steps
\FOR {$i$ in $I$}
    \STATE translated $\gets LLM(i)$
    \STATE translated $\gets$ PRUNE(translated)
\ENDFOR
\STATE $S \gets \{\}$
\FOR {$i$ in translated}
    \FOR {$t_i$ in $i$}
        \STATE $S[t_i] \gets SIGN\_VIDEO(t_i)$ 
\ENDFOR
\ENDFOR
\STATE $V \gets [ \ ]$
\FOR {$i$ in $I$}
    \FOR {$t_i$ in $i$}
        \STATE $V[i] \gets V[i] + S[t_i]$
\ENDFOR
\ENDFOR \\
\RETURN $V$
\end{algorithmic}
\end{algorithm}

\subsection{Large Language Model Translation}
\label{sec:llm}
For the translation of spoken English instructions to textual representations of ASL (glosses), we prompt LLMs. Multiple methods exist in implementing text-to-gloss translation: human annotation, rule-based automatic translation with heuristics \cite{Othman2012}, fine-tuned transformer-based models \cite{Camgoz2018, yin-read-2020-better}, and prompting LLMs \cite{Lee}. We make our system adaptable to all of these alternatives for text-to-gloss translation. Any one of these models can be plugged into line 4 of Algorithm~\ref{alg:sign}. 
We choose LLM translation for our current system due to its scalability, translation understandability, and ability to adapt to out-of-domain text. We show in our system evaluation in section \S \ref{sec:eval} that there is a trade-off between using LLMs or rule-based heuristics for text-to-gloss translation. Mainly, LLMs generate more diverse translations, while rule-based heuristics have higher accuracy depending on the video dataset size.


Our instructions consist of WholeFoods recipes\footnote{\url{www.wholefoodsmarket.com/recipes}} and WikiHow tasks\footnote{\url{www.wikihow.com}}. First, we aggregate all the instruction steps of the task in a JSON construct (given in Appendix~\ref{sec:inputs}), then using the OpenAI chat API we prompt \texttt{gpt-3.5-turbo} to ``translate each step to American Sign Language gloss", and request the result in a JSON format.\footnote{Our parameters for the API call are, \textit{temperature}=1, max \textit{tokens}=1000, \textit{top p}=1, \textit{frequency penalty}=0, and \textit{presence penalty}=0.} We then aggregate all these steps for all recipes and tasks. For recipes, we do not translate the ingredients to glosses, as our community outreach surveys indicate that users prefer to see the ingredients written statically on the screen instead of signed versions (see Figure~\ref{fig:sign_pipeline} for a reference of text-to-gloss translation steps). 

After these instructions are generated, we have an additional stage of manual correction of LLM-generated glosses using rule-based heuristics for quality\footnote{this curation step can be omitted for the deployment of larger systems with bigger task sets, where it might be infeasible to go over each task step and glosses manually.}. We also remove the punctuation in glosses, capitalize them, and concatenate the fingerspellings---in which fingers form individual letters to spell out words---if annotated using the hyphen notation (i.e. ``F-I-N-G-E-R"). Here, we check that the glosses are unique across the tasks, they are all present in the available video dictionary, and they follow the general rules of ASL. 

\vspace{-5pt}

\subsection{Sign Video Processing}
We process the videos in four steps. First, we collect sign videos corresponding to all the glosses in our instruction set from an online platform. Then we store these videos, retrieve them on the fly while presenting instructions, and stitch them together. We give the details of these steps in the following paragraphs.

\paragraph{Sign Video Collection} For video collection, we use widely available American Sign Language sign dictionary videos from video sharing platforms with Creative Commons licenses online \footnote{\url{Lifeprint.com}, and the ASLDictionary channel accessible on YouTube: \url{https://youtube.com/@smartsigndictionary}}. We mainly use videos from Lifeprint, but if they do not contain a specific sign video, we use the ASLDictionary on YouTube as the backup source. If neither of these sources has a sign available, we first check if the gloss can be deconstructed into other signs or fingerspelled. If so, we check the videos for the deconstructed versions and concatenate them into a single video. If these options are not available and the gloss is crucial to the meaning of the instruction, then we search for a synonym. If it is not crucial to the meaning of the instruction, then we drop the gloss.

\paragraph{Video Storage} We generate a dictionary for all the available sign glosses (found in Appendix Section~\ref{sec:inputs}) and upload all the videos with their gloss as their filename to an Amazon AWS S3 bucket for storage. 

\paragraph{Gloss-by-Gloss Sign Retrieval} During a user's live use of the system for signed instructions, we retrieve videos on a token level using the video URL by cross-referencing its gloss filename. As the last step, after retrieving all the video URLs on the fly for each gloss in each instruction, we concatenate all of the URLs corresponding to the glosses together and then present them on the user interface of the app as a single stream of a video (see Figure~\ref{fig:sign_demo}).

\section{System Evaluation}
\label{sec:eval}
We evaluate our system both quantitatively and qualitatively. Because this is the first deployment of a task-oriented signed multimodal dialogue system, we chiefly compare the system with the non-signed portion of our task-oriented dialogue system. We first evaluate the performance of our LLM text-to-gloss translation and discuss the trade-offs of using an LLM for translation. Then, we evaluate our algorithm using traditional information retrieval metrics. Finally, we compare user ratings and provide detailed qualitative analyses by an expert who is fluent in ASL. 

\paragraph{Text-to-Gloss Translation Analysis}
\begin{table}[!h]
    \centering
    \resizebox{\columnwidth}{!}{
    \begin{tabular}{cccccccc}
        \toprule
        \multicolumn{8}{c}{\textbf{Automatic Metrics}} \\
        \midrule
        \multicolumn{4}{c}{\textbf{BLEU}} & \multirow{2}{*}{\textbf{ROUGE}} & \multirow{2}{*}{\textbf{METEOR}} & \multirow{2}{*}{\textbf{ChrF}} & \multirow{2}{*}{\textbf{WER}}\\
        \textbf{1} & \textbf{2} & \textbf{3} & \textbf{4} & & & & \\ 
        \midrule
        9.52 & 1.59 & 0.42 & 0.16 & 0.11 & 0.11 & 23.99 & 2.146 \\
        \toprule
        & & \multicolumn{2}{c}{\textbf{F1}} & \multicolumn{2}{c}{\textbf{Recall}} & \multicolumn{2}{c}{\textbf{Precision}} \\
        \cmidrule{3-8}
        \multicolumn{2}{c|}{\textbf{BERTScore}} & \multicolumn{2}{c}{$0.80$} & \multicolumn{2}{c}{$0.81$} & \multicolumn{2}{c}{$0.79$} \\
        \bottomrule
    \end{tabular}
    }
    \caption{This table shows the automatic metric results between LLM and rule-based translations. Tasks on the web do not contain readily available ground-truth glosses. BERTScore is the best indicator of translation success.}
    \label{tab:llm_results}
\end{table}

In this section, we analyze the performance of LLM-based translations using traditional automatic text metrics (see Table~\ref{tab:llm_results}). As also described in section \S~\ref{sec:llm}, we experiment with two translation strategies: 1) LLM translations and 2) rule-based gloss translations with heuristics. We use the rule-based heuristics strategy as ground truth in our results here because no human-annotated ASL ground truth exists for our datasets, and the accuracy of rule-based translations is high when compared to human annotations in the works of \citet{Othman2012EnglishASLGP, Othman2019Apr}. In order to generate rule-based glosses, we use the Algorithm given in Appendix \ref{sec:rule_based_algo}. Automatic evaluation metrics for sign translations do not yet exist. Hence, we present results using traditional automatic evaluation metrics such as BLEU \cite{papineni-etal-2002-bleu}, ROUGE \cite{lin-2004-rouge}, METEOR \cite{banerjee-lavie-2005-meteor}, ChrF \cite{popovic-2015-chrf}, and BERTScore \cite{zhang2020bertscore} between LLM-generated glosses and the rule-based glosses. 
In this case, BERTScore is more insightful than traditional metrics because the semantic representation of tokens is more important in glossing than the specific n-gram differences. 

For our system, we deploy with LLM-based translations and are able to scale from only 1-3 tasks with ASL expert manual annotations to 150 supported tasks with LLM-based translations. As shown in Figure~\ref{fig:dialogue_sys}, the LLM translations happen offline as all of our tasks are pre-determined. Right after the tasks are translated to ASL glosses, we have a quality control stage before they are presented to the user. So, our overall translation pipeline is a human-in-the-loop system. During the duration of our dialogue system's deployment, we observe that using LLMs reduces the time spent on the manual checking process by human annotators from 10 minutes per instruction sentence to 1 minute per sentence.

\begin{figure}[!h]
    \centering
    \begin{tabular}{cc}
        \includegraphics[width=\columnwidth]{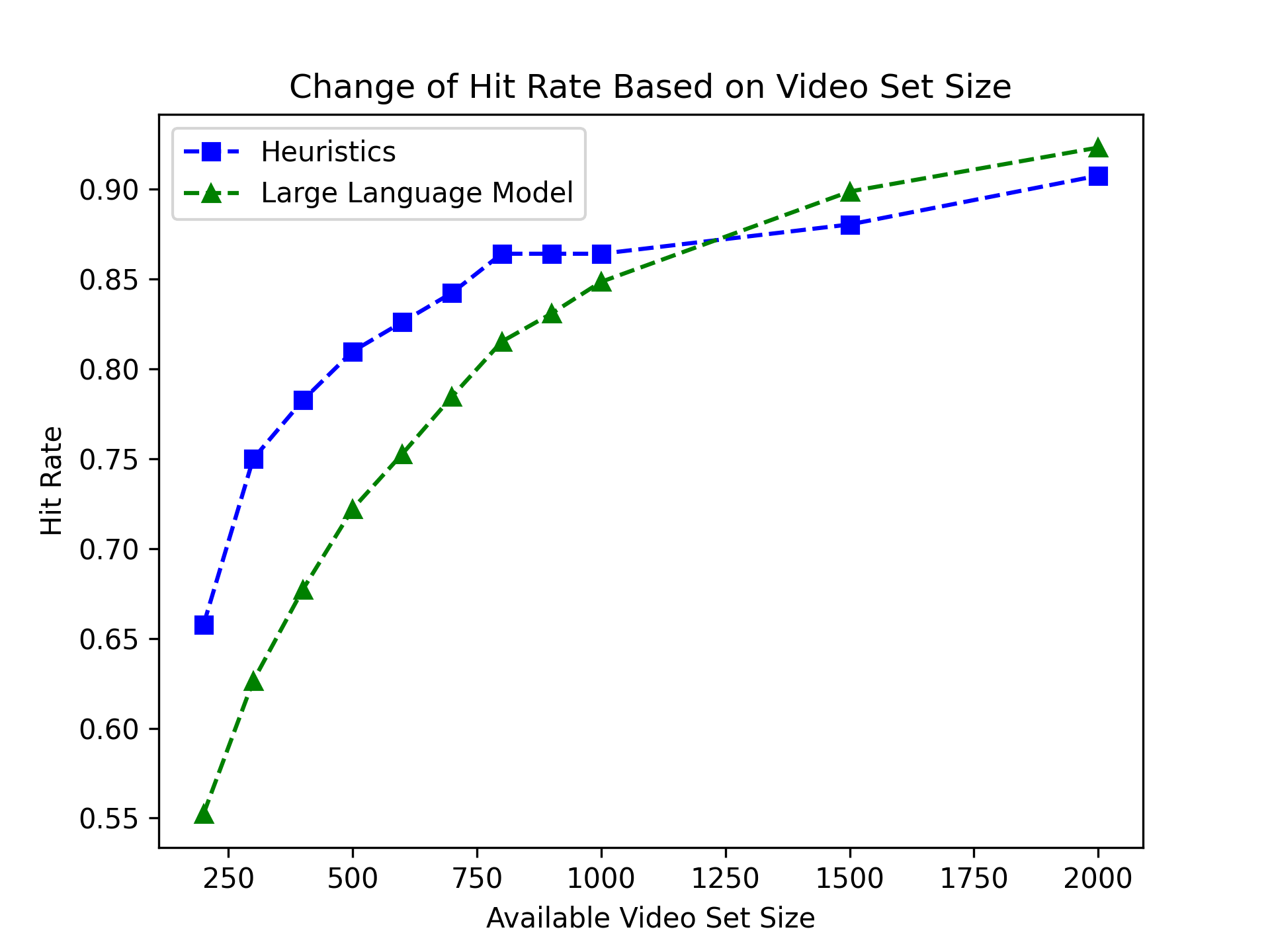} \\
        \includegraphics[width=\columnwidth]{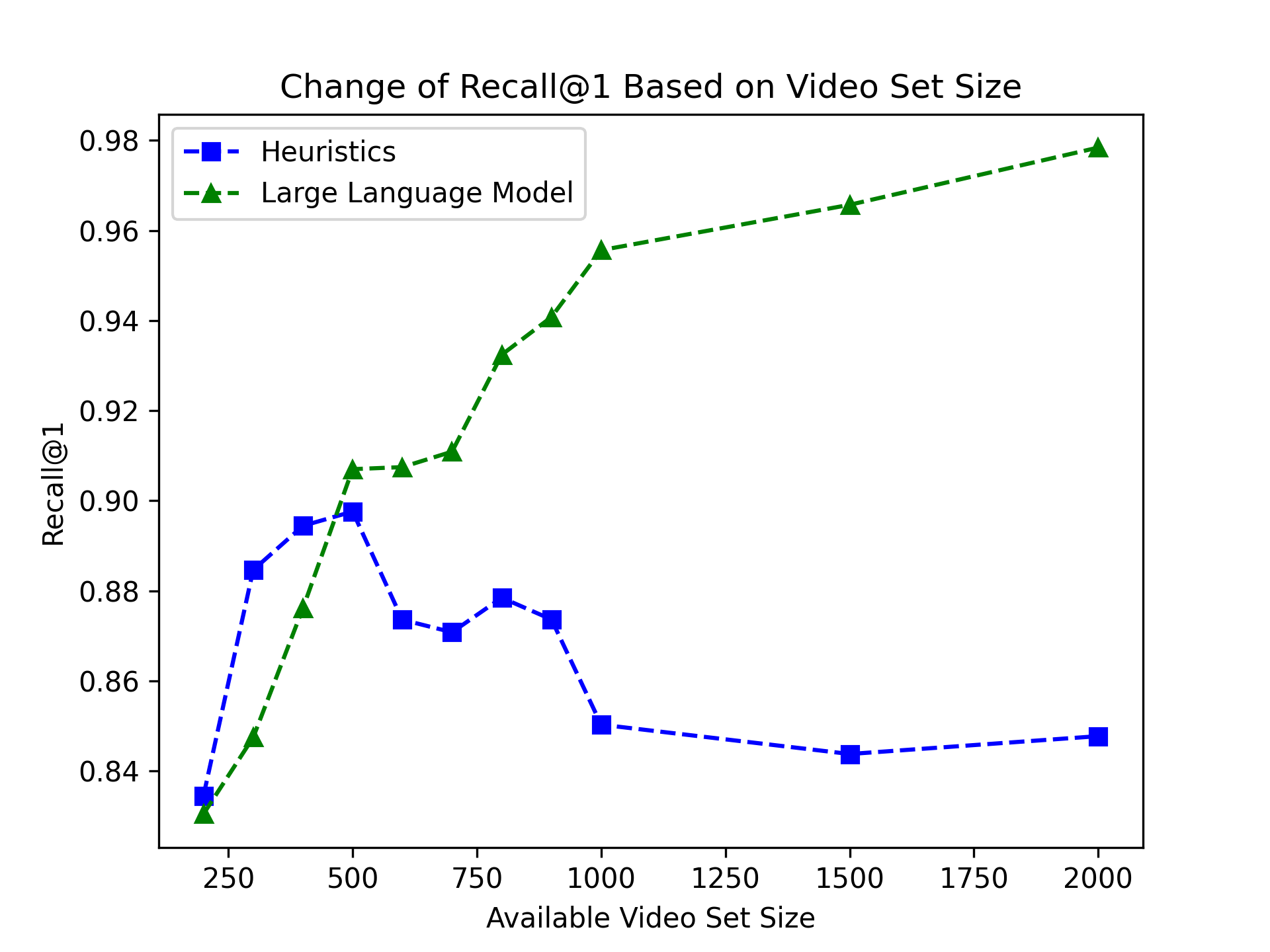}
    \end{tabular}
        
    \caption{These plots show the changes in Hit Rate and Recall@1 for our signed instruction retrieval algorithm as the available video set increases in size. Two lines represent two methods of translation from text to gloss. In a constrained setup with limited sign video storage, these plots show how many videos are needed with different translation strategies. Overall, LLMs have more diverse translations, while rule-based heuristics provide more accurate translations changing with the video dataset size.}
    \label{fig:retrieval_metrics}
\end{figure}

\paragraph{Retrieval Metrics}
No automatic evaluation mechanism exists for signed interactive systems; hence, in this section, we introduce two retrieval metrics---Hit Rate and Recall@1---for our Signed Instruction Retrieval Algorithm (see Algorithm~\ref{alg:sign}) with the two translation modules separately. Furthermore, we also present an analysis of the changes in Hit Rate and Recall@1 in response to increases in the available video dataset size in Figure~\ref{fig:retrieval_metrics}.

We use the following simplified definitions of Hit Rate and Recall@1:
\begin{align}
    \text{Hit Rate} &= \frac{\#~\text{glosses w/ videos}}{\text{total}~\#~\text{of glosses}} \\
    \text{Recall}@1 &= \frac{\#~\text{glosses w/ videos}}{\begin{aligned}\#~\text{synonyms of glosses w/o videos} \\ ~+~ \# ~\text{glosses w/ videos} \end{aligned}}
\end{align}



Essentially, Hit Rate measures how accurate the system is in finding videos for a given token, and Recall@1 tells how precise the system selects videos corresponding to a token among a set of synonyms. For instance, for a task step consisting of glosses ``\texttt{CHOCOLATE CHOP ADD DOUGH MIX STIR}'' if the system has only videos for \texttt{CHOP}, \texttt{ADD}, \texttt{COMBINE}, and \texttt{STIR}, then the Hit Rate will be $0.5$, as three out of six glosses do not have videos; and Recall@1 will be $3/4$, where the denominator also contains any synonym of a gloss that does not have a corresponding video (\texttt{MIX} and \texttt{COMBINE} are considered synonyms in this case). Hit Rate and Recall@1 are complimentary metrics where Hit Rate shows the direct presence of sign videos while Recall@1 indirectly shows how diverse the glosses and selected videos are due to the inclusion of synonyms in the denominator where multiple glosses may exist for the same video that we have in our database. We give detailed mathematical definitions for both of these metrics in Appendix~\ref{sec:metrics}.

Looking at the resulting plots in Figure~\ref{fig:retrieval_metrics}, we can make several claims. For Hit Rate, both of the translation strategies produce similar results because our video database covers a majority of glosses present in the restricted domain of cooking and wikiHow tasks. For Recall@1, there is a dramatic difference between LLMs and heuristics. This happens because rule-based heuristics use nearly the same tokens from the text, while LLMs can generate synonymous glosses for a given token. For a more example-driven explanation, please refer to Appendix~\ref{sec:ret_met_examples}.


Overall, the Recall@1 for our Algorithm has a minimum of around 80\% and a maximum of 98\%---as observed in Figure~\ref{fig:retrieval_metrics}. This shows that our algorithm can easily be deployed as part of dialogue systems with signed instructions regardless of whether we use LLMs or rule-based heuristics translations.




\paragraph{User Rating Comparisons}
Our system interacted with a large number of public users for over a period of six months. Because this is the first task-oriented dialogue system with signed instructions, it increases our user outreach on international platforms by a large margin. However, adding this functionality could decrease overall user ratings if they do not deem the interface usable or are unsure about what ASL is. Thus, we examine the ratings before and after adding the signed instructions to our system. As shown in Appendix~\ref{fig:user-ratings}, our user ratings remain constant after adding support for this feature. Thus, we find that, besides making task-oriented systems accessible to a larger audience, adding support for signed instructions does not decrease user ratings.

\paragraph{Expert Qualitative Analysis}

One author fluent in ASL evaluated the system with special regard to the usability and clarity of the information presented. This evaluator noted \textit{two primary strengths}: 1) the multimodal instructional support provided by having both the ASL descriptions and the instructional images available, particularly for the step-by-step tasks such as origami folding; 2) the ease of processing and attending to multiple modalities given the clear layout without overwhelming the user. To expand, giving the user the option to attend to the signed content or the referent of the images (e.g., step-by-step origami folding) allowed them to rely on each form of information to the extent they prefer. The clear layout does not overwhelm the user with too many streams of information. It also allows for sufficient processing of either sign videos, images, or both without distracting the user. 

The \textit{primary limitation} of the current system lies in the segmented nature of the ASL videos. Currently, there is a lack of smooth transitions between signs, and different signers present each sign within one instruction. The flow of the signs appears disjointed, consequently impeding clear understanding. 
The absence of step-by-step visuals in certain tasks necessitates increased reliance on signing. The disjointed nature of the current signing videos rendered some tasks less comprehensible. 

Overall, the multimodal presentation of signing alongside informative images enhances accessibility and suggests that a dynamic display of signed content will greatly enhance future task-oriented dialogue systems. For future iterations of our system, we plan to incorporate either human models signing the entire content or synthesized avatars \cite{Quandt2020Oct, Quandt2022Feb}.  





\section{Conclusion}
In this work, we discussed a multimodal, task-oriented dialogue system designed to generate ASL instructions on a platform with global reach. Emphasizing the critical importance of Deaf and Hard-of-Hearing (DHH) community engagement throughout the development cycle, our approach integrates extensive feedback from both the signing community and experts in the field. Our system not only marks a significant technological milestone but also enriches the dialogue on how video-based ASL instruction delivery can be effectively scaled internationally. We observed a nuanced preference among signers for avatar-based instructions---a finding underscored by our expert analysis. Our system has improved the landscape of conversational AI, making it accessible and responsive to the unique needs of the DHH community.

 We make the code available for our pipeline and encourage future researchers to incorporate it into their own task-oriented systems to increase accessibility. We hope that this system is a step towards developing dialogue systems that can understand \emph{and} generate signs for all signed languages.   
 We encourage everybody to interact with signed tasks by visiting \url{https://huggingface.co/spaces/merterm/signed-instructions}.

\section{Acknowledgement}
 This project was completed as part of and received funding from the Alexa Prize TaskBot Challenge 2. We would like to thank the Alexa Prize team, especially Lavina Vaz and Michael Johnston, for supporting us throughout the competition and for giving us the resources to develop and deploy our system to a large audience. We would also like to thank our team members: Yuya Asano, Qi Cheng, Dipunj Gupta, Sabit Hassan, Jennifer Nwogu, and Paras Sharma.

\bibliography{custom}

\begin{thebibliography}{40}
\expandafter\ifx\csname natexlab\endcsname\relax\def\natexlab#1{#1}\fi

\bibitem[{Agichtein et~al.(2023)Agichtein, Johnston, Gottardi, Flagg, Vaz, Shi, Zhang, Ball, Liu, Dai, Pressel, Goyal, Hu, Ipek, Sahai, Lu, Liu, Hakkani-Tür, Hu, Rocker, Jeun, Iyengar, Mandal, Kuzi, Vedula, Rokhlenko, Castellucci, Choi, Bland, , Maarek, and Ghanadan}]{TaskBot2023}
Eugene Agichtein, Michael Johnston, Anna Gottardi, Cris Flagg, Lavina Vaz, Hangjie Shi, Desheng Zhang, Leslie Ball, Shaohua Liu, Luke Dai, Daniel Pressel, Prasoon Goyal, Lucy Hu, Osman Ipek, Sattvik Sahai, Yao Lu, Yang Liu, Dilek Hakkani-Tür, Shui Hu, Heather Rocker, James Jeun, Akshaya Iyengar, Arindam Mandal, Saar Kuzi, Nikhita Vedula, Oleg Rokhlenko, Giuseppe Castellucci, Jason~Ingyu Choi, Kate Bland, , Yoelle Maarek, and Reza Ghanadan. 2023.
\newblock \href {https://www.amazon.science/alexa-prize/proceedings/alexa-lets-work-together-introducing-the-second-alexa-prize-taskbot-challenge} {Alexa, let’s work together: Introducing the second alexa prize taskbot challenge}.
\newblock In \emph{Alexa Prize TaskBot Challenge 2 Proceedings}.

\bibitem[{Anindhita and Lestari(2016)}]{Anindhita2016Aug}
Vidia Anindhita and Dessi~Puji Lestari. 2016.
\newblock \href {https://doi.org/10.1109/ICAICTA.2016.7803135} {{Designing interaction for deaf youths by using user-centered design approach}}.
\newblock In \emph{{2016 International Conference On Advanced Informatics: Concepts, Theory And Application (ICAICTA)}}, pages 1--6. IEEE.

\bibitem[{Banerjee and Lavie(2005)}]{banerjee-lavie-2005-meteor}
Satanjeev Banerjee and Alon Lavie. 2005.
\newblock \href {https://aclanthology.org/W05-0909} {{METEOR}: An automatic metric for {MT} evaluation with improved correlation with human judgments}.
\newblock In \emph{Proceedings of the {ACL} Workshop on Intrinsic and Extrinsic Evaluation Measures for Machine Translation and/or Summarization}, pages 65--72, Ann Arbor, Michigan. Association for Computational Linguistics.

\bibitem[{Blair and Abdullah(2020)}]{Blair2020Dec}
Johnna Blair and Saeed Abdullah. 2020.
\newblock \href {https://doi.org/10.1145/3432194} {{It Didn't Sound Good with My Cochlear Implants: Understanding the Challenges of Using Smart Assistants for Deaf and Hard of Hearing Users}}.
\newblock \emph{Proc. ACM Interact. Mob. Wearable Ubiquitous Technol.}, 4(4):1--27.

\bibitem[{Bragg et~al.(2019)Bragg, Koller, Bellard, Berke, Boudreault, Braffort, Caselli, Huenerfauth, Kacorri, Verhoef, Vogler, and Ringel~Morris}]{Bragg2019Oct}
Danielle Bragg, Oscar Koller, Mary Bellard, Larwan Berke, Patrick Boudreault, Annelies Braffort, Naomi Caselli, Matt Huenerfauth, Hernisa Kacorri, Tessa Verhoef, Christian Vogler, and Meredith Ringel~Morris. 2019.
\newblock \href {https://doi.org/10.1145/3308561.3353774} {{Sign Language Recognition, Generation, and Translation: An Interdisciplinary Perspective}}.
\newblock In \emph{{ASSETS '19: Proceedings of the 21st International ACM SIGACCESS Conference on Computers and Accessibility}}, pages 16--31. Association for Computing Machinery, New York, NY, USA.

\bibitem[{Bragg et~al.(2020)Bragg, Morris, Vogler, Kushalnagar, Huenerfauth, and Kacorri}]{Bragg2020Apr}
Danielle Bragg, Meredith~Ringel Morris, Christian Vogler, Raja Kushalnagar, Matt Huenerfauth, and Hernisa Kacorri. 2020.
\newblock \href {https://doi.org/10.1145/3334480.3381053} {{Sign Language Interfaces: Discussing the Field's Biggest Challenges}}.
\newblock In \emph{{CHI EA '20: Extended Abstracts of the 2020 CHI Conference on Human Factors in Computing Systems}}, pages 1--5. Association for Computing Machinery, New York, NY, USA.

\bibitem[{Camgoz et~al.(2018)Camgoz, Hadfield, Koller, Ney, and Bowden}]{Camgoz2018}
Necati~Cihan Camgoz, Simon Hadfield, Oscar Koller, Hermann Ney, and Richard Bowden. 2018.
\newblock \href {https://openaccess.thecvf.com/content_cvpr_2018/html/Camgoz_Neural_Sign_Language_CVPR_2018_paper.html} {{Neural Sign Language Translation}}.
\newblock [Online; accessed 9. Oct. 2023].

\bibitem[{Chambers(2020)}]{Chambers2020}
Cindy Chambers. 2020.
\newblock \href {https://digitalcommons.wou.edu/theses/139} {{Mindfulness and Interpreter Cognitive Load}}.
\newblock \emph{Digital Commons@WOU}.

\bibitem[{Dangsaart et~al.(2008)Dangsaart, Naruedomkul, Cercone, and Sirinaovakul}]{Dangsaart2008Nov}
Srisavakon Dangsaart, Kanlaya Naruedomkul, Nick Cercone, and Booncharoen Sirinaovakul. 2008.
\newblock \href {https://doi.org/10.1016/j.compedu.2007.11.008} {{Intelligent Thai text {\textendash} Thai sign translation for language learning}}.
\newblock \emph{Computers {\&} Education}, 51(3):1125--1141.

\bibitem[{Du~Toit(2017)}]{DuToit2017}
P.~T.~Petri Du~Toit. 2017.
\newblock \href {https://wiredspace.wits.ac.za/items/9adcf705-637c-4fc9-909e-7779dbef53e0} {{Mitigating the cognitive load of South African Sign Language interpreters on national television}}.
\newblock [Online; accessed 20. Jul. 2023].

\bibitem[{Glasser et~al.(2017)Glasser, Kushalnagar, and Kushalnagar}]{Glasser2017Oct}
Abraham Glasser, Kesavan Kushalnagar, and Raja Kushalnagar. 2017.
\newblock \href {https://doi.org/10.1145/3132525.3134781} {{Deaf, Hard of Hearing, and Hearing Perspectives on Using Automatic Speech Recognition in Conversation}}.
\newblock In \emph{{ASSETS '17: Proceedings of the 19th International ACM SIGACCESS Conference on Computers and Accessibility}}, pages 427--432. Association for Computing Machinery, New York, NY, USA.

\bibitem[{Glasser et~al.(2020)Glasser, Mande, and Huenerfauth}]{Glasser2020Jul}
Abraham Glasser, Vaishnavi Mande, and Matt Huenerfauth. 2020.
\newblock \href {https://doi.org/10.1145/3405755.3406158} {{Accessibility for Deaf and Hard of Hearing Users: Sign Language Conversational User Interfaces}}.
\newblock In \emph{{CUI '20: Proceedings of the 2nd Conference on Conversational User Interfaces}}, pages 1--3. Association for Computing Machinery, New York, NY, USA.

\bibitem[{Hariharan et~al.(2018)Hariharan, Al-khazraji, and Huenerfauth}]{Hariharan2018Jun}
Dhananjai Hariharan, Sedeeq Al-khazraji, and Matt Huenerfauth. 2018.
\newblock \href {https://doi.org/10.1007/978-3-319-92049-8_15} {{Evaluation of an English Word Look-Up Tool for Web-Browsing with Sign Language Video for Deaf Readers}}.
\newblock In \emph{{Universal Access in Human-Computer Interaction. Methods, Technologies, and Users}}, pages 205--215. Springer, Cham, Switzerland.

\bibitem[{Hr{\ifmmode\acute{u}\else\'{u}\fi}z et~al.(2011)Hr{\ifmmode\acute{u}\else\'{u}\fi}z, Campr, Kr{\ifmmode\check{n}\else\v{n}\fi}oul, {\ifmmode\check{Z}\else\v{Z}\fi}elezn{\ifmmode\acute{y}\else\'{y}\fi}, Aran, and Santemiz}]{Hruz2011Oct}
Marek Hr{\ifmmode\acute{u}\else\'{u}\fi}z, Pavel Campr, Zdenek Kr{\ifmmode\check{n}\else\v{n}\fi}oul, Milos {\ifmmode\check{Z}\else\v{Z}\fi}elezn{\ifmmode\acute{y}\else\'{y}\fi}, Oya Aran, and Pinar Santemiz. 2011.
\newblock \href {https://doi.org/10.1145/2049536.2049599} {{Multi-modal dialogue system with sign language capabilities}}.
\newblock In \emph{{ASSETS '11: The proceedings of the 13th international ACM SIGACCESS conference on Computers and accessibility}}, pages 265--266. Association for Computing Machinery, New York, NY, USA.

\bibitem[{Huang et~al.()Huang, Wu, and Kameda}]{Huang}
Xuan Huang, Bo~Wu, and Hiroyuki Kameda.
\newblock \href {https://doi.org/10.1109/DASC-PICom-CBDCom-CyberSciTech52372.2021.00144} {{Development of a Sign Language Dialogue System for a Healing Dialogue Robot}}.
\newblock In \emph{{2021 IEEE Intl Conf on Dependable, Autonomic and Secure Computing, Intl Conf on Pervasive Intelligence and Computing, Intl Conf on Cloud and Big Data Computing, Intl Conf on Cyber Science and Technology Congress (DASC/PiCom/CBDCom/CyberSciTech)}}, pages 25--28. IEEE.

\bibitem[{Inan et~al.(2022)Inan, Zhong, Hassan, Quandt, and Alikhani}]{inan-etal-2022-modeling}
Mert Inan, Yang Zhong, Sabit Hassan, Lorna Quandt, and Malihe Alikhani. 2022.
\newblock \href {https://doi.org/10.18653/v1/2022.findings-acl.228} {Modeling intensification for sign language generation: A computational approach}.
\newblock In \emph{Findings of the Association for Computational Linguistics: ACL 2022}, pages 2897--2911, Dublin, Ireland. Association for Computational Linguistics.

\bibitem[{Kahlon and Singh(2023)}]{Kahlon2023Mar}
Navroz~Kaur Kahlon and Williamjeet Singh. 2023.
\newblock \href {https://doi.org/10.1007/s10209-021-00823-1} {{Machine translation from text to sign language: a systematic review}}.
\newblock \emph{Univ. Access Inf. Soc.}, 22(1):1--35.

\bibitem[{Kelly(2003)}]{kelly2003considerations}
Leonard~P Kelly. 2003.
\newblock Considerations for designing practice for deaf readers.
\newblock \emph{Journal of deaf studies and deaf education}, 8(2):171--186.

\bibitem[{Lee et~al.()Lee, Kim, Hwang, Kim, and Park}]{Lee}
Huije Lee, Jung-Ho Kim, Eui~Jun Hwang, Jaewoo Kim, and Jong~C. Park.
\newblock \href {https://doi.org/10.1109/ICASSPW59220.2023.10193533} {{Leveraging Large Language Models With Vocabulary Sharing For Sign Language Translation}}.
\newblock In \emph{{2023 IEEE International Conference on Acoustics, Speech, and Signal Processing Workshops (ICASSPW)}}, pages 04--10. IEEE.

\bibitem[{Lin(2004)}]{lin-2004-rouge}
Chin-Yew Lin. 2004.
\newblock \href {https://aclanthology.org/W04-1013} {{ROUGE}: A package for automatic evaluation of summaries}.
\newblock In \emph{Text Summarization Branches Out}, pages 74--81, Barcelona, Spain. Association for Computational Linguistics.

\bibitem[{Lin et~al.(2023)Lin, Wang, Zhu, Sun, Zhang, and Yang}]{lin-etal-2023-gloss}
Kezhou Lin, Xiaohan Wang, Linchao Zhu, Ke~Sun, Bang Zhang, and Yi~Yang. 2023.
\newblock \href {https://doi.org/10.18653/v1/2023.acl-long.722} {Gloss-free end-to-end sign language translation}.
\newblock In \emph{Proceedings of the 61st Annual Meeting of the Association for Computational Linguistics (Volume 1: Long Papers)}, pages 12904--12916, Toronto, Canada. Association for Computational Linguistics.

\bibitem[{Luckner and Handley(2008)}]{luckner2008summary}
John~L Luckner and C~Michele Handley. 2008.
\newblock A summary of the reading comprehension research undertaken with students who are deaf or hard of hearing.
\newblock \emph{American annals of the deaf}, 153(1):6--36.

\bibitem[{Macnamara(2012)}]{Macnamara2012}
Brooke Macnamara. 2012.
\newblock \href {https://digitalcommons.unf.edu/joi/vol19/iss1/1/?utm_source=digitalcommons.unf.edu%2Fjoi%2Fvol19%2Fiss1%2F1&utm_medium=PDF&utm_campaign=PDFCoverPages} {{Interpreter Cognitive Aptitudes}}.
\newblock \emph{Journal of Interpretation}, 19(1):1.

\bibitem[{Moryossef et~al.(2021)Moryossef, Yin, Neubig, and Goldberg}]{moryossef-etal-2021-data}
Amit Moryossef, Kayo Yin, Graham Neubig, and Yoav Goldberg. 2021.
\newblock \href {https://aclanthology.org/2021.mtsummit-at4ssl.1} {Data augmentation for sign language gloss translation}.
\newblock In \emph{Proceedings of the 1st International Workshop on Automatic Translation for Signed and Spoken Languages (AT4SSL)}, pages 1--11, Virtual. Association for Machine Translation in the Americas.

\bibitem[{M{\"u}ller et~al.(2023)M{\"u}ller, Jiang, Moryossef, Rios, and Ebling}]{muller-etal-2023-considerations}
Mathias M{\"u}ller, Zifan Jiang, Amit Moryossef, Annette Rios, and Sarah Ebling. 2023.
\newblock \href {https://doi.org/10.18653/v1/2023.acl-short.60} {Considerations for meaningful sign language machine translation based on glosses}.
\newblock In \emph{Proceedings of the 61st Annual Meeting of the Association for Computational Linguistics (Volume 2: Short Papers)}, pages 682--693, Toronto, Canada. Association for Computational Linguistics.

\bibitem[{Nasihati~Gilani et~al.(2019)Nasihati~Gilani, Traum, Sortino, Gallagher, Aaron-Lozano, Padilla, Shapiro, Lamberton, and Petitto}]{NasihatiGilani2019Jul}
Setareh Nasihati~Gilani, David Traum, Rachel Sortino, Grady Gallagher, Kailyn Aaron-Lozano, Cryss Padilla, Ari Shapiro, Jason Lamberton, and Laura-Ann Petitto. 2019.
\newblock \href {https://doi.org/10.1145/3308532.3329463} {{Can a Signing Virtual Human Engage a Baby's Attention?}}
\newblock In \emph{{IVA '19: Proceedings of the 19th ACM International Conference on Intelligent Virtual Agents}}, pages 162--169. Association for Computing Machinery, New York, NY, USA.

\bibitem[{Othman and Jemni(2012{\natexlab{a}})}]{Othman2012}
Achraf Othman and M.~Jemni. 2012{\natexlab{a}}.
\newblock \href {https://www.semanticscholar.org/paper/English-ASL-Gloss-Parallel-Corpus-2012%3A-ASLG-PC12-Othman-Jemni/473fffb95c3db24938a21346ecd117a8a9204404?p2df} {{English-ASL Gloss Parallel Corpus 2012: ASLG-PC12}}.
\newblock [Online; accessed 20. Jul. 2023].

\bibitem[{Othman and Jemni(2012{\natexlab{b}})}]{Othman2012EnglishASLGP}
Achraf Othman and Mohamed Jemni. 2012{\natexlab{b}}.
\newblock \href {https://api.semanticscholar.org/CorpusID:67028968} {English-asl gloss parallel corpus 2012: Aslg-pc12}.

\bibitem[{Othman and Jemni(2019)}]{Othman2019Apr}
Achraf Othman and Mohamed Jemni. 2019.
\newblock \href {https://doi.org/10.4018/JITR.2019040108} {{Designing High Accuracy Statistical Machine Translation for Sign Language Using Parallel Corpus: Case Study English and American Sign Language}}.
\newblock \emph{J. Inf. Technol. Res.}, 12(2):134--158.

\bibitem[{Papineni et~al.(2002)Papineni, Roukos, Ward, and Zhu}]{papineni-etal-2002-bleu}
Kishore Papineni, Salim Roukos, Todd Ward, and Wei-Jing Zhu. 2002.
\newblock \href {https://doi.org/10.3115/1073083.1073135} {{B}leu: a method for automatic evaluation of machine translation}.
\newblock In \emph{Proceedings of the 40th Annual Meeting of the Association for Computational Linguistics}, pages 311--318, Philadelphia, Pennsylvania, USA. Association for Computational Linguistics.

\bibitem[{Popovi{\'c}(2015)}]{popovic-2015-chrf}
Maja Popovi{\'c}. 2015.
\newblock \href {https://doi.org/10.18653/v1/W15-3049} {chr{F}: character n-gram {F}-score for automatic {MT} evaluation}.
\newblock In \emph{Proceedings of the Tenth Workshop on Statistical Machine Translation}, pages 392--395, Lisbon, Portugal. Association for Computational Linguistics.

\bibitem[{Quandt(2020)}]{Quandt2020Oct}
Lorna Quandt. 2020.
\newblock \href {https://doi.org/10.1145/3373625.3418042} {{Teaching ASL Signs using Signing Avatars and Immersive Learning in Virtual Reality}}.
\newblock In \emph{{ASSETS '20: Proceedings of the 22nd International ACM SIGACCESS Conference on Computers and Accessibility}}, pages 1--4. Association for Computing Machinery, New York, NY, USA.

\bibitem[{Quandt et~al.(2022)Quandt, Willis, Schwenk, Weeks, and Ferster}]{Quandt2022Feb}
Lorna~C. Quandt, Athena Willis, Melody Schwenk, Kaitlyn Weeks, and Ruthie Ferster. 2022.
\newblock \href {https://doi.org/10.3389/fpsyg.2022.730917} {{Attitudes Toward Signing Avatars Vary Depending on Hearing Status, Age of Signed Language Acquisition, and Avatar Type}}.
\newblock \emph{Front. Psychol.}, 13:730917.

\bibitem[{Tiselius(2018)}]{Tiselius2018Jun}
Elisabet Tiselius. 2018.
\newblock \href {https://doi.org/10.7146/hjlcb.v0i57.106193} {{Exploring Cognitive Aspects of Competence in Sign Language Interpreting of Dialogues: First Impressions}}.
\newblock \emph{HJLCB}, (57):49--61.

\bibitem[{Traxler(2000)}]{Traxler2000Sep}
Carol~Bloomquist Traxler. 2000.
\newblock \href {https://doi.org/10.1093/deafed/5.4.337} {{The Stanford Achievement Test, 9th Edition: National Norming and Performance Standards for Deaf and Hard-of-Hearing Students}}.
\newblock \emph{J. Deaf Stud. Deaf Educ.}, 5(4):337--348.

\bibitem[{Viegas et~al.(2023)Viegas, Inan, Quandt, and Alikhani}]{viegas-etal-2023-including}
Carla Viegas, Mert Inan, Lorna Quandt, and Malihe Alikhani. 2023.
\newblock \href {https://doi.org/10.18653/v1/2023.starsem-1.1} {Including facial expressions in contextual embeddings for sign language generation}.
\newblock In \emph{Proceedings of the 12th Joint Conference on Lexical and Computational Semantics (*SEM 2023)}, pages 1--10, Toronto, Canada. Association for Computational Linguistics.

\bibitem[{Wojtanowski et~al.(2020)Wojtanowski, Gilmore, Seravalli, Fargas, Vogler, and Kushalnagar}]{Wojtanowski2020}
Gabriella Wojtanowski, Colleen Gilmore, Barbra Seravalli, Kristen Fargas, Christian Vogler, and Raja Kushalnagar. 2020.
\newblock \href {https://scholarworks.csun.edu/handle/10211.3/215984} {{"Alexa, Can You See Me?" Making Individual Personal Assistants for the Home Accessible to Deaf Consumers}}.
\newblock \emph{California State University, Northridge.}

\bibitem[{Yin et~al.(2021)Yin, Moryossef, Hochgesang, Goldberg, and Alikhani}]{yin-etal-2021-including}
Kayo Yin, Amit Moryossef, Julie Hochgesang, Yoav Goldberg, and Malihe Alikhani. 2021.
\newblock \href {https://doi.org/10.18653/v1/2021.acl-long.570} {Including signed languages in natural language processing}.
\newblock In \emph{Proceedings of the 59th Annual Meeting of the Association for Computational Linguistics and the 11th International Joint Conference on Natural Language Processing (Volume 1: Long Papers)}, pages 7347--7360, Online. Association for Computational Linguistics.

\bibitem[{Yin and Read(2020)}]{yin-read-2020-better}
Kayo Yin and Jesse Read. 2020.
\newblock \href {https://doi.org/10.18653/v1/2020.coling-main.525} {Better sign language translation with {STMC}-transformer}.
\newblock In \emph{Proceedings of the 28th International Conference on Computational Linguistics}, pages 5975--5989, Barcelona, Spain (Online). International Committee on Computational Linguistics.

\bibitem[{Zhang et~al.(2020)Zhang, Kishore, Wu, Weinberger, and Artzi}]{zhang2020bertscore}
Tianyi Zhang, Varsha Kishore, Felix Wu, Kilian~Q. Weinberger, and Yoav Artzi. 2020.
\newblock \href {http://arxiv.org/abs/1904.09675} {Bertscore: Evaluating text generation with bert}.

\end{thebibliography}

\appendix

\clearpage
\onecolumn
\section{Input Constructs}
\label{sec:inputs}
Here we show the JSON format of the tasks:
\begin{figure*}[hbt!]
    \includegraphics[width=\textwidth]{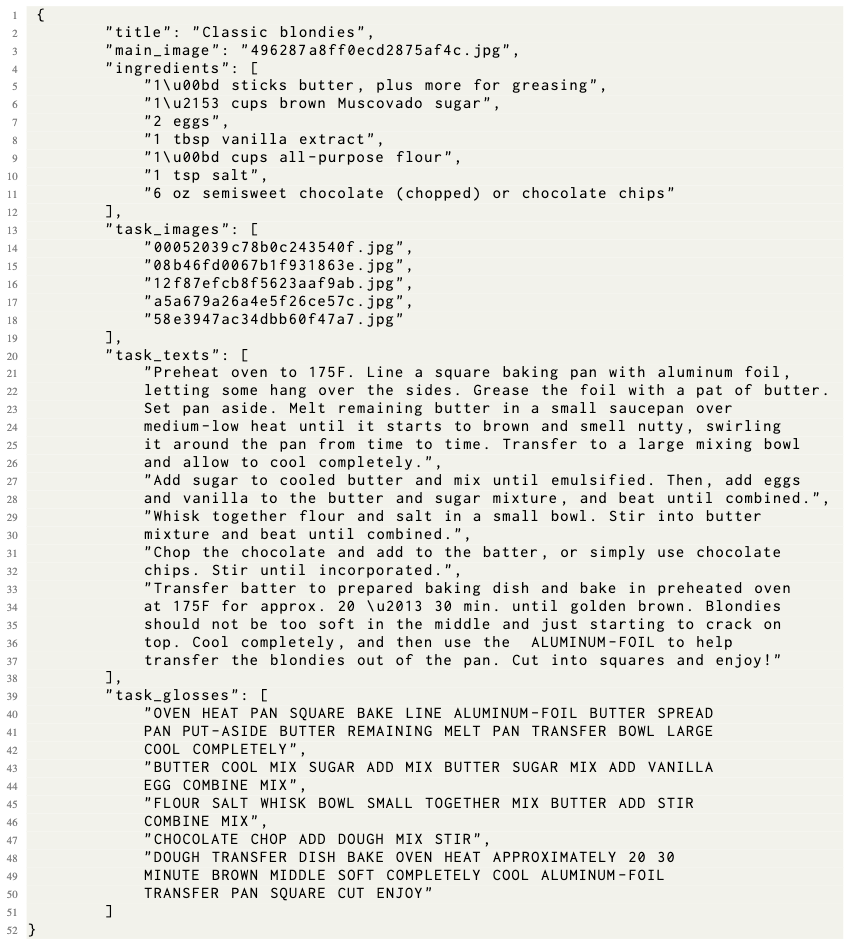}
    \label{code:json}
\end{figure*}

\clearpage

Here is the dictionary of all the available glosses that have corresponding videos on the system.
\begin{figure*}[!h]
    \includegraphics[width=\textwidth]{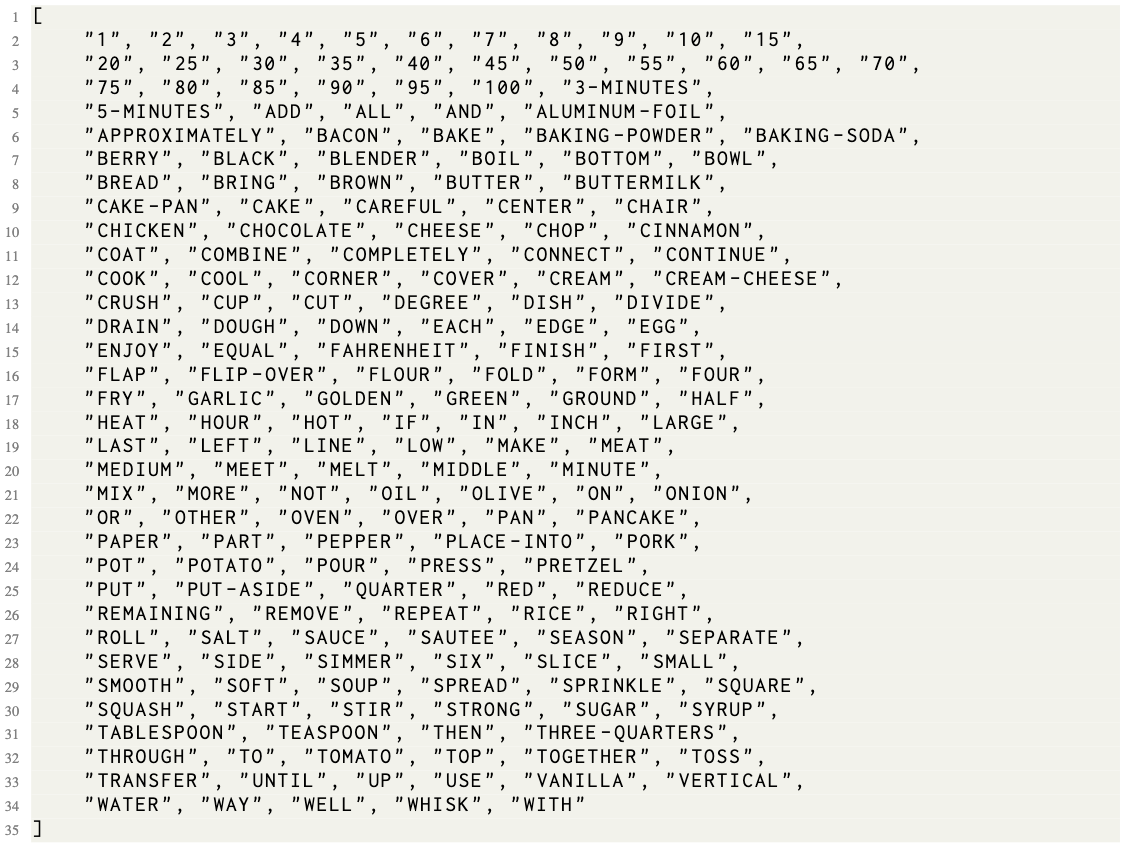}
    \label{code:gloss_dict}
\end{figure*}

\clearpage
\twocolumn

\section{Rule-based Gloss Translation Algorithm}
\label{sec:rule_based_algo}
We give the pseudocode for the rule-based heuristics algorithm as follows:

\begin{algorithm}[!h]
\begin{algorithmic}[1]
\caption{Rule-based Heuristic Glosses}\label{alg:heuristics}
\STATE $heuristic\_glosses \gets []$
\FOR {$sentence$ \textbf{in} $task['task\_texts']$}
    \STATE $sentence \gets$ \textsc{UpperCase}($sentence$)
    \STATE $text \gets$ \textsc{Tokenize}($sentence$)
    \STATE $pos\_tagged \gets$ \textsc{POStagging}($text$)
    \FOR {$token$ \textbf{in} $pos\_tagged$}
        \IF {\textit{IsNotDesiredPOS}($token[1]$)}
            \STATE \textsc{RemoveToken}($pos\_tagged, token$)
        \ENDIF
    \ENDFOR
    \FOR {$i$ \textbf{in range}(\textsc{Length}($pos\_tagged$))}
        \STATE $pos\_tagged[i] \gets$ (\textsc{Lemmatize}($pos\_tagged[i][0]$), $pos\_tagged[i][1]$)
    \ENDFOR
    \STATE $sentence \gets$ ""
    \FOR {$token$ \textbf{in} $pos\_tagged$}
        \STATE $sentence \gets sentence + token[0] + " "$
    \ENDFOR
    \STATE $sentence \gets$ \textsc{Strip}($sentence$)
    \STATE $heuristic\_glosses$.APPEND($sentence$)
\ENDFOR
\RETURN $heuristic\_glosses$
\end{algorithmic}
\end{algorithm}

\section{Detailed Mathematical Definitions for Retrieval Metrics}
\label{sec:metrics}
To define Hit Rate and Recall@1 more precisely, we first introduce some requisite definitions: 
\begin{itemize}
\setlength\itemsep{-.5em}
  \item $D$: set of glosses in our dictionary
  \item $n$: total number of task instructions 
  \item $I=\{i_0, i_1, ..., i_n\}$: set of all task instructions 
  \item $m_k$: total number of glosses in instruction $k$ 
  \item $i_k\in I = <g_{k0}, g_{k1},..., g_{km_k}>$  
  \item $g_{kl} \in i_k$: gloss in instruction $i_k$ (ordered) 
  \item $syn(g)$: the set of synonyms found for gloss $g$ using {\tt wordnet.synsets}
\end{itemize}


We formalize our simplified definitions of Hit Rate and Recall@1 below, using our notation. Note that because we take into account repeated glosses in our instruction set, the sets below are \emph{multisets} and thus contain repeated elements that are factored into the cardinality of the set.

\begin{align}
    \text{Hit Rate} &= \frac{|{g_{kl}:g_{kl}\in D,i_k\in I, g_{kl} \in i_k}|}{|{g_{kl}:i_k\in I, g_{kl} \in i_k}|} \\
    \text{Recall}@1 &= \frac{|{g_{kl}:g_{kl}\in D,i_k\in I, g_{kl} \in i_k}|}{\begin{aligned}|{g_{kl}:g_{kl}\in D,i_k\in I, g_{kl} \in i_k}|\\~+~|{g_{kl}:g_{kl}\notin D,i_k\in I, g_{kl} \in i_k}|\end{aligned}}
\end{align}

\section{Detailed Examples for Retrieval Metrics}
\label{sec:ret_met_examples}
For example, for the instruction, \textit{``Chop chocolate and add to batter. Stir until incorporated.''}, the LLM generates, ``\texttt{CHOCOLATE CHOP ADD DOUGH MIX STIR}'', while heuristics generates ``\texttt{CHOP CHOCOLATE ADD BATTER STIR UNTIL INCORPORATE}''. Here, it can be seen that LLM produces \texttt{DOUGH} (a synonym of ``batter'' for our purposes), while heuristics directly uses the same wording. This adds diversity to the generated glosses, and as the number of videos increases, it positively affects the score of LLMs. For the heuristics algorithm, as the tokens are never changed into synonyms, even after a lot of videos are added to the set, the algorithm cannot retrieve videos and gets lower Recall@1 scores.


        

\clearpage
\onecolumn
\section{Interface Details}
\label{sec:interface}
\begin{figure*}[!h]
\centering
    \resizebox{\linewidth}{!}{
    \begin{tabular}{ccc}
        \includegraphics[width=.33\linewidth]{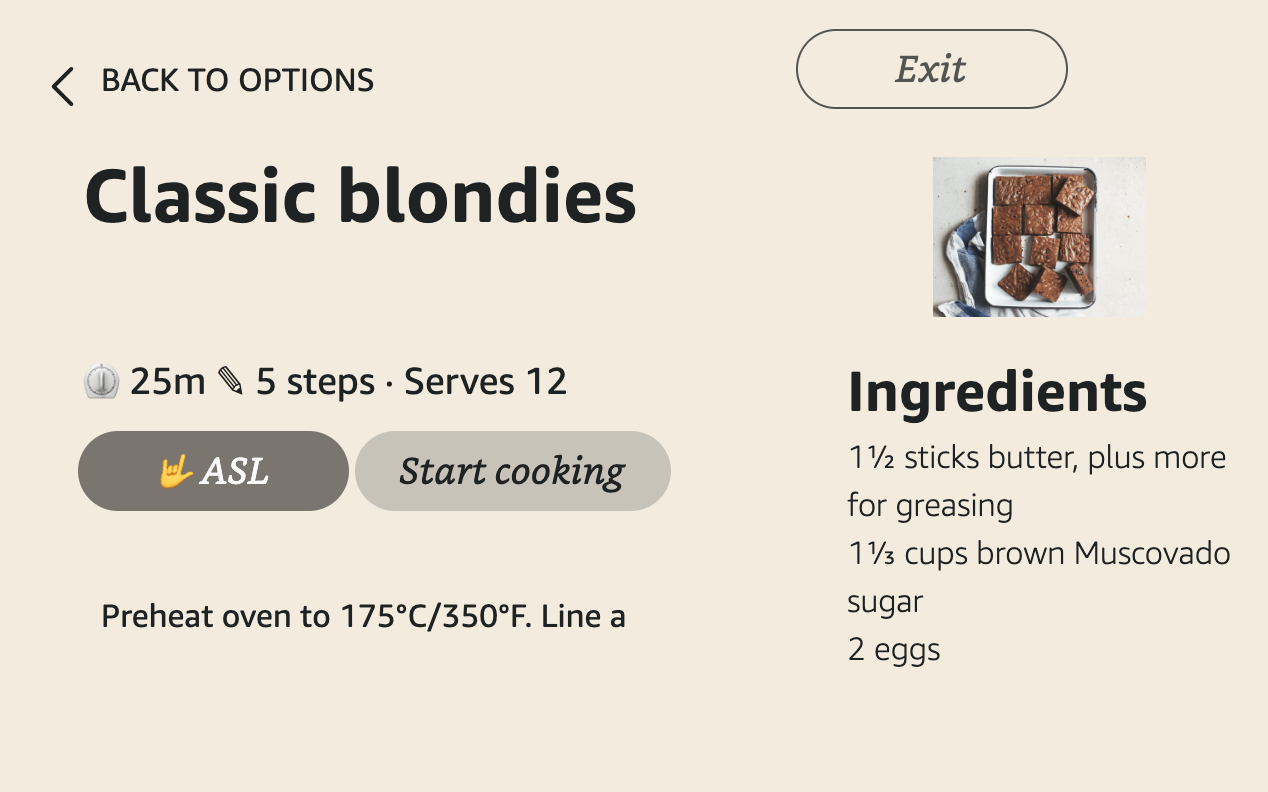} & \includegraphics[width=.33\linewidth]{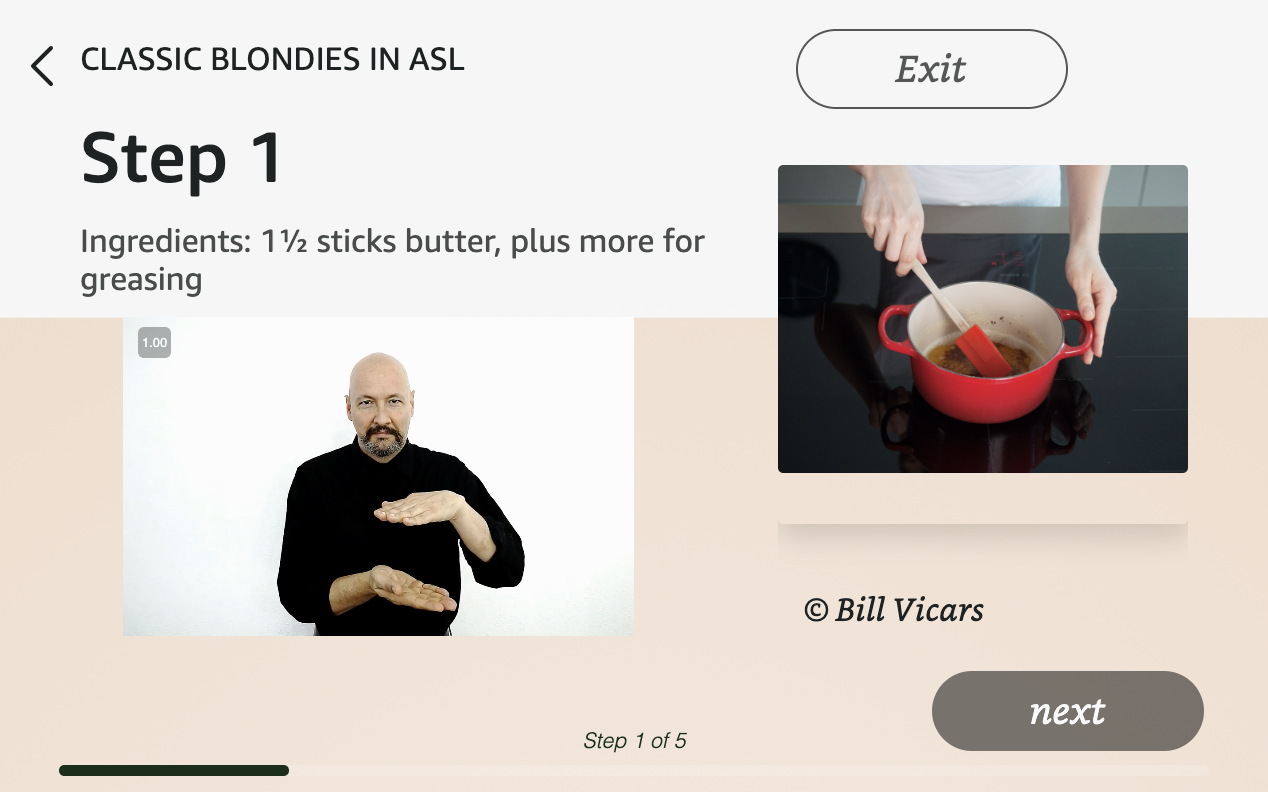} &
        \includegraphics[width=.33\linewidth]{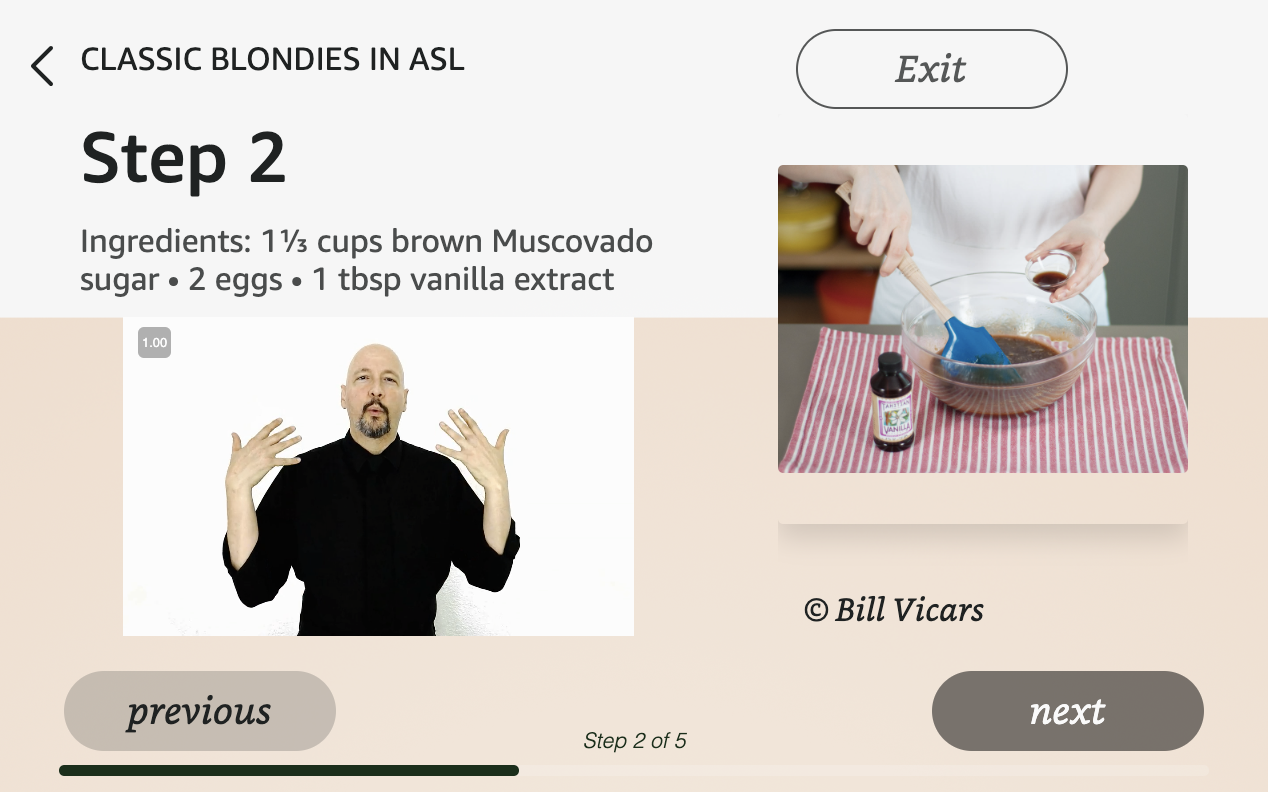}
    \end{tabular}
    }
    \caption{These are the screens for an alternative task of a classic blondies recipe. The main difference for recipes is that at each step, relevant ingredients are shown in addition to the signed instruction video. This is to ensure less cognitive load on the user. Also, the first panel shows the ASL button that exists in supported recipes.}
    \label{fig:sign_demo_detail1}
\end{figure*}

\begin{figure*}[!h]
    \centering
    \resizebox{\linewidth}{!}{
    \begin{tabular}{cc}
        \includegraphics[width=.5\linewidth]{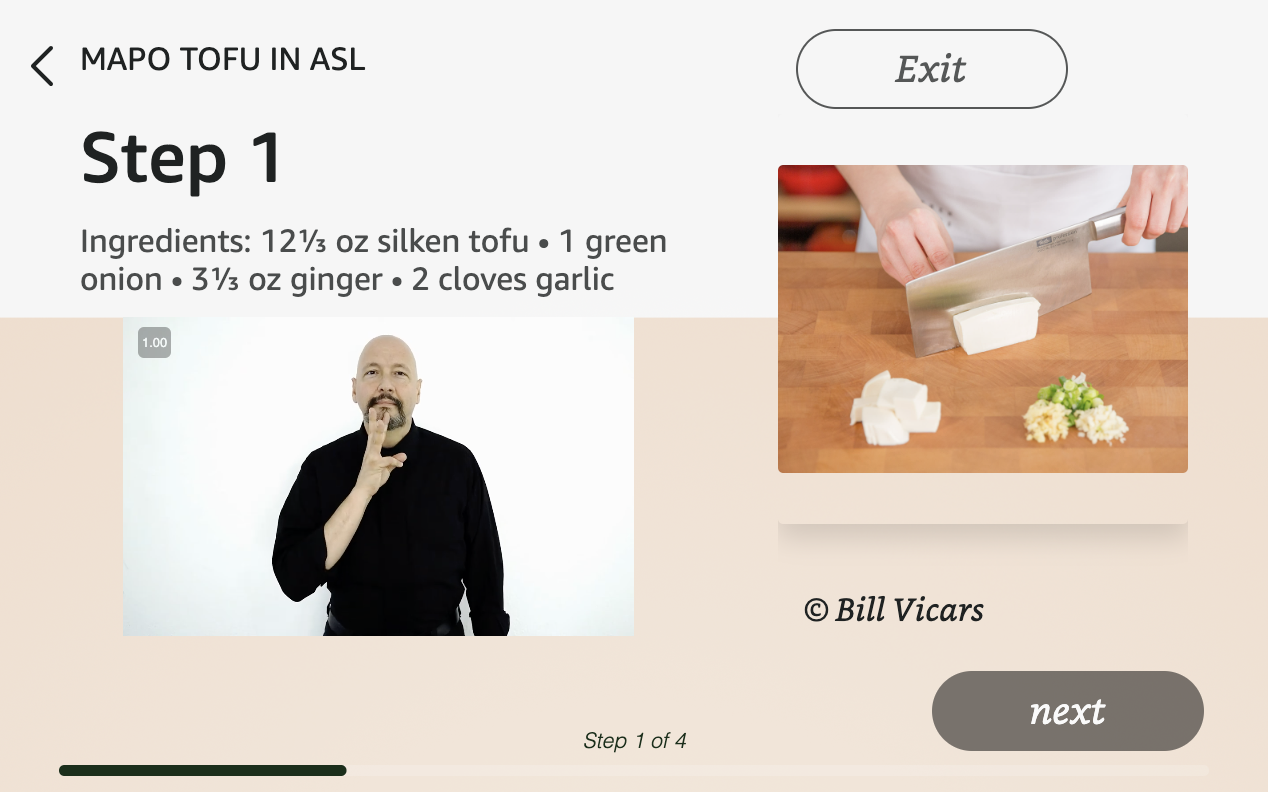} & \includegraphics[width=.5\linewidth]{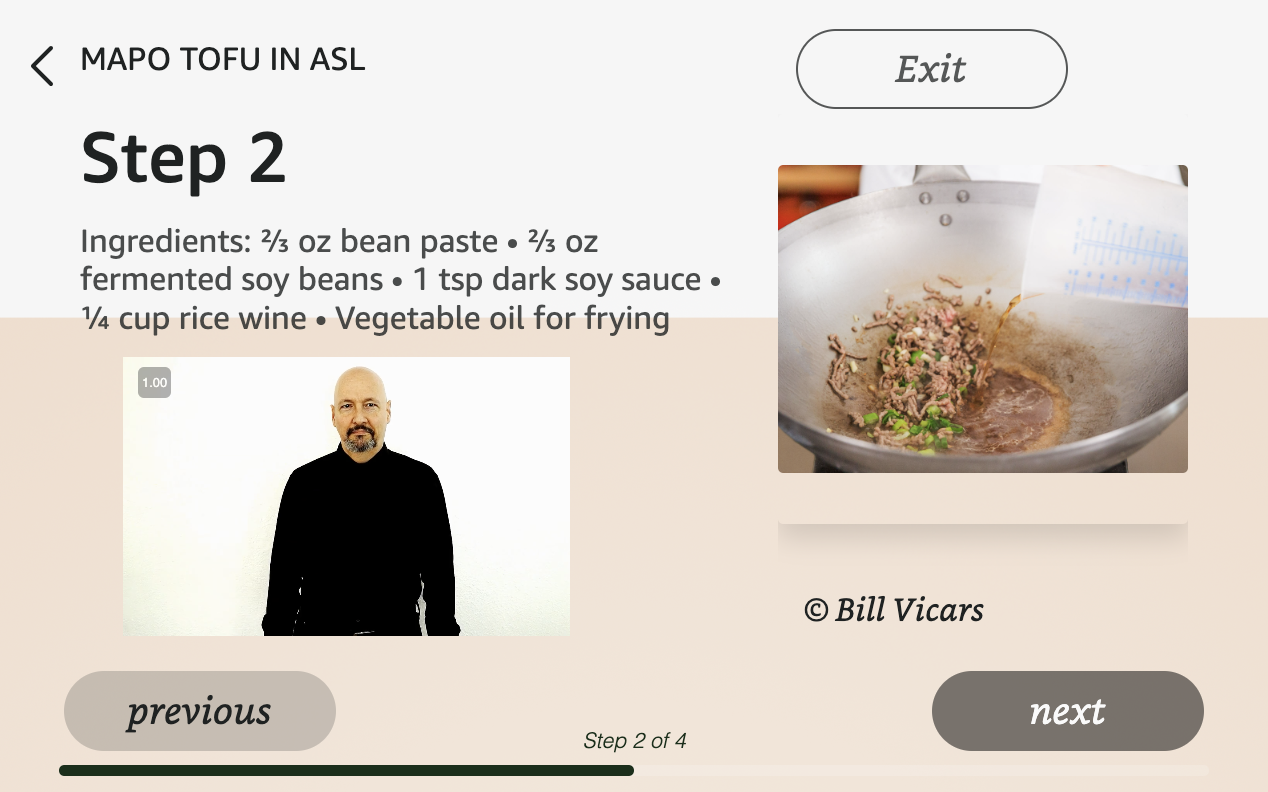} \\
        \includegraphics[width=.5\linewidth]{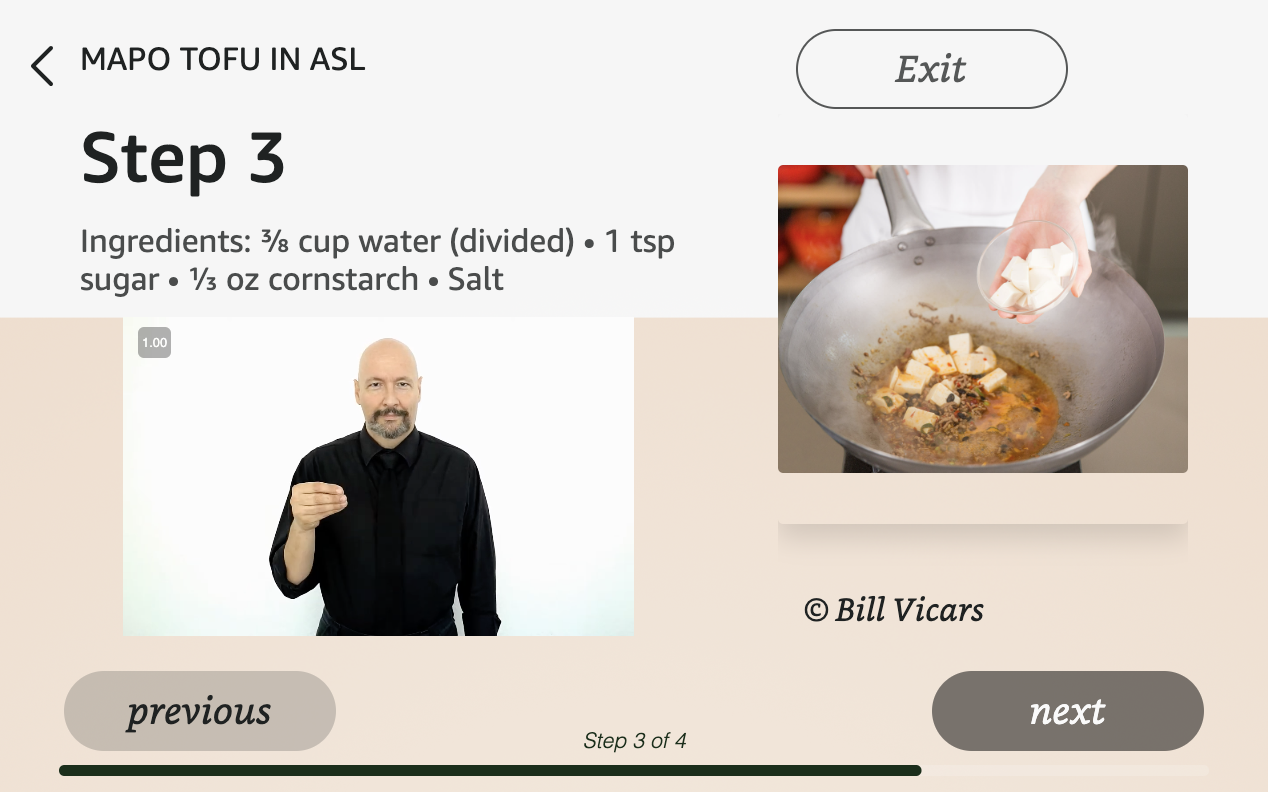} & \includegraphics[width=.5\linewidth]{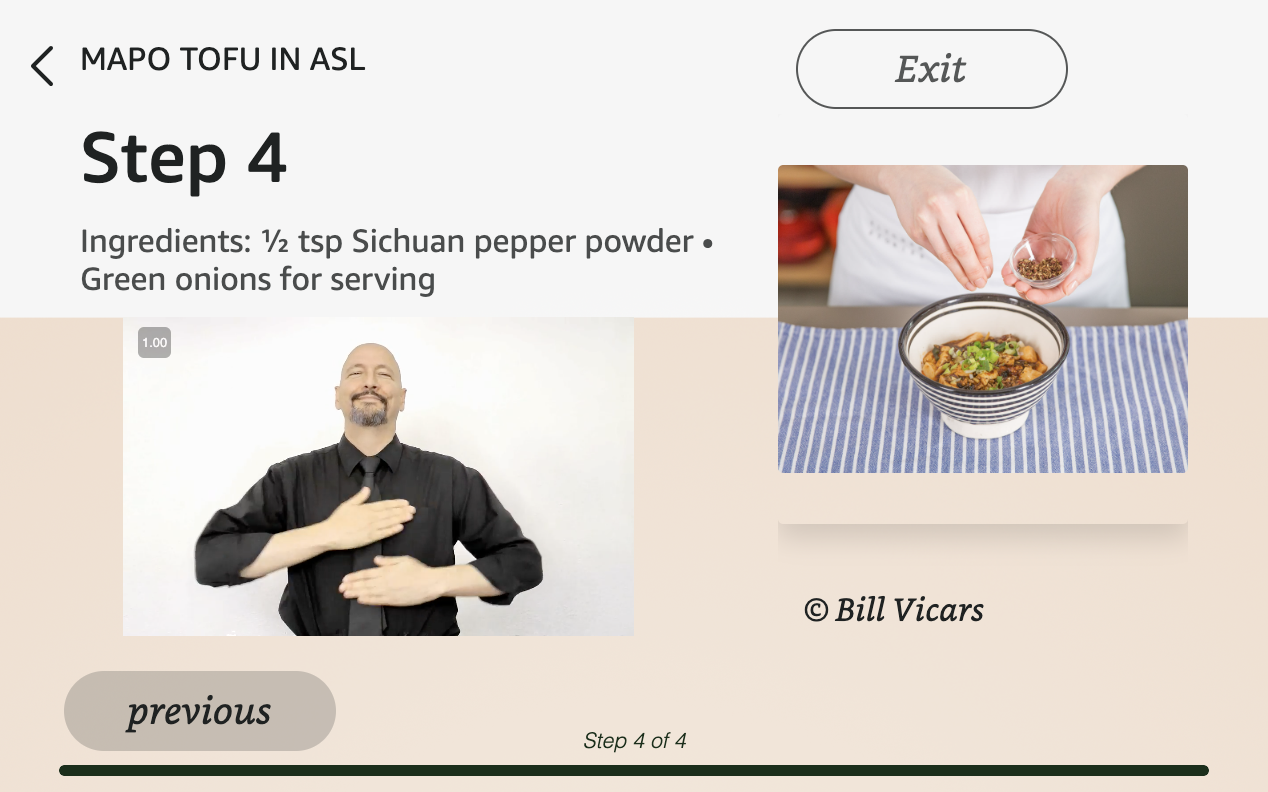}
    \end{tabular}
    }
    \caption{This figure demonstrates the screenshots of our signed multimodal dialogue bot for the recipe of Mapo Tofu. This example is chosen to stress the fact that certain international recipes that have terms that may not exist in ASL are also supported in the bot. In these cases, the ingredients are written on the screen and the instructions are signed without the specific terminologies, like "tofu", and images are shown to aid with grounding the referred ingredient.}
    \label{fig:sign_demo_detail2}
\end{figure*}

We show more screenshots of details in the interface in Figures~\ref{fig:sign_demo_detail1}, and~\ref{fig:sign_demo_detail2}.

\clearpage
\section{User Rating Analysis}
\begin{figure*}[!h]
    \centering
    \includegraphics[width=\textwidth]{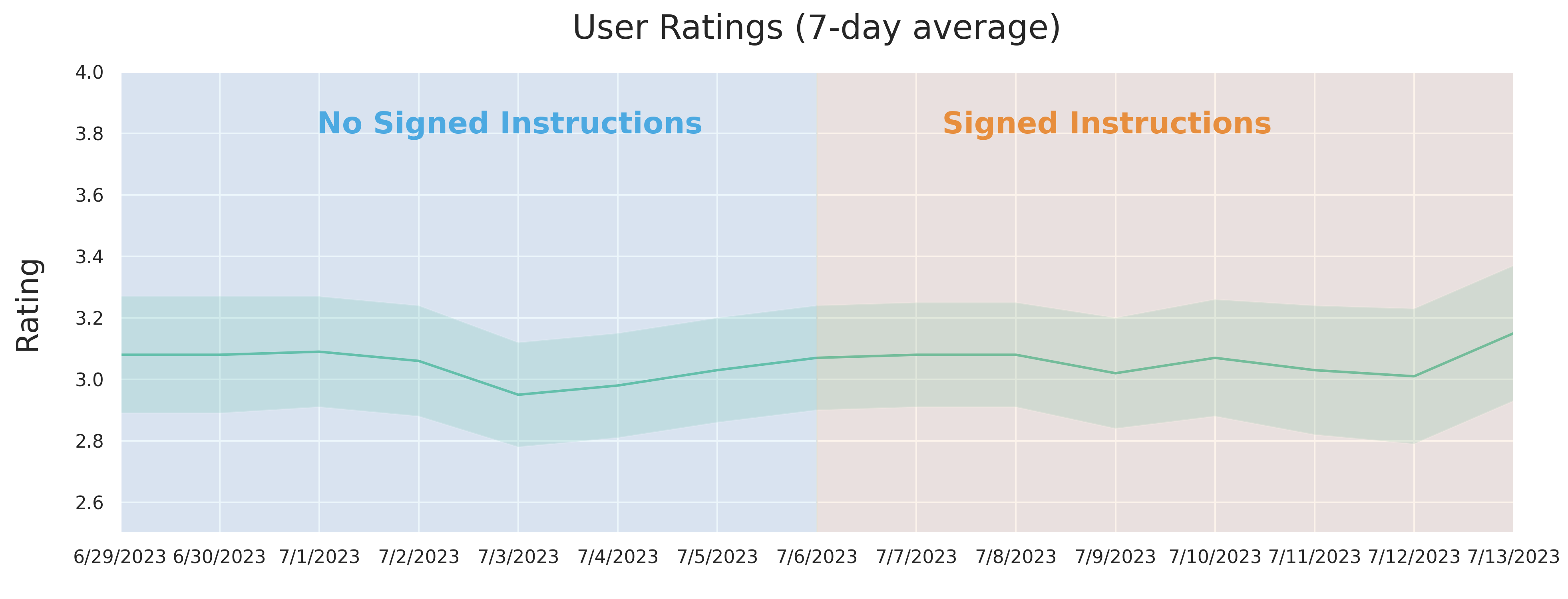}
    \caption{User ratings of our system before and after adding support for instructions in ASL.  Here, we show the week before and after adding signed instructions. Reaching out to real users and communities that use signed languages is the main goal of our system. Adding ASL support allows our system to engage with a larger audience without decreasing overall user ratings.} 
    \label{fig:user-ratings}
\end{figure*}

We show a plot of 7-day averages of user ratings before and after adding support for signed instructions in Figure~\ref{fig:user-ratings}.

\end{document}